\documentclass{article}

\PassOptionsToPackage{numbers, compress}{natbib}

\usepackage[preprint]{neurips_data_2023}

\usepackage{CJKutf8}
\usepackage{lipsum}
\usepackage[table]{xcolor}
\usepackage{tcolorbox}

\usepackage{url}            
\usepackage{booktabs}       
\usepackage{amsfonts}       
\usepackage{amsmath}
\usepackage{hyperref}
\usepackage{graphicx,calc} 
\usepackage{makecell}
\usepackage{wrapfig}

\newcommand{\Rmnum}[1]{\expandafter\@slowromancap\romannumeral #1@}
\usepackage{bbm}
\usepackage{amssymb}
\usepackage{multirow}
\usepackage{fontawesome5}
\usepackage{nicematrix} 
\usepackage{paralist}
\usepackage[edges]{forest}
\usepackage{siunitx}
\usepackage{tikz-dependency}
\usepackage{tikz}
\usepackage{pifont}
\usepackage{cleveref}

\newlength\myheight
\newlength\mydepth
\settototalheight\myheight{Xygp}
\definecolor{uclablue}{rgb}{0.15, 0.45, 0.68}
\definecolor{lightcoral}{rgb}{0.94, 0.5, 0.5}
\definecolor{lightgreen}{rgb}{0.56, 0.93, 0.56}
\definecolor{harvestgold}{rgb}{0.85, 0.57, 0.0}
\definecolor{brightlavender}{rgb}{0.75, 0.58, 0.89}
\definecolor{capri}{rgb}{0, 0.61, 0.94}
\definecolor{carminepink}{rgb}{0.92, 0.3, 0.26}
\definecolor{celadon}{rgb}{0.67, 0.88, 0.69}
\definecolor{darkpastelgreen}{rgb}{0.01, 0.75, 0.24}
\definecolor{cplus}{rgb}{1,0.42,0}
\definecolor{cminus}{rgb}{0,0.48,0.76}
\definecolor{tablegrey}{HTML}{EFEFEF}

\crefname{figure}{Figure}{Figures}
\crefname{table}{Table}{Tables} 
\crefname{section}{Section}{Sections}

\definecolor{headcolor}{RGB}{218,232,252}

\newcommand\refsec[1]{\hyperref[sec:#1]{§\ref{sec:#1}:~\textsc{#1}}}
\newcommand\refsecs[2]{\hyperref[sec:#1]{§\ref{sec:#1}:~\textsc{#1}}, \hyperref[sec:#2]{§\ref{sec:#2}:~\textsc{#2}}}

\providecommand{\keywords}[1]
{
  \small
  \textbf{\textit{Keywords---}} #1
}

\hypersetup{
    breaklinks,
    citecolor=uclablue,
    colorlinks=true,
    linkcolor=uclablue
}

\newcommand{\eg}{\emph{e.g.}}
\newcommand{\ie}{\emph{i.e.}}

\newcommand*\colourcheck[1]{%
  \expandafter\newcommand\csname #1check\endcsname{\textcolor{#1}{\ding{52}}}%
}
\newcommand*\colourcross[1]{%
  \expandafter\newcommand\csname #1cross\endcsname{\textcolor{#1}{\ding{55}}}%
}

\colourcheck{blue}
\colourcheck{green}
\colourcheck{darkpastelgreen}
\colourcross{blue}
\colourcross{applegreen}
\colourcross{red}

\title{Multimodal Chain-of-Thought Reasoning:\\ A Comprehensive Survey}

\author{
Yaoting Wang$^1$, \, 
Shengqiong Wu$^1$, \, 
Yuechen Zhang$^2$, \\
\textbf{Shuicheng Yan$^1$, \, 
Ziwei Liu$^3$,  \, 
Jiebo Luo$^4$, \, 
Hao Fei$^1$\thanks{Corresponding Author. (\texttt{haofei37@nus.edu.sg})}
}
\\
{$^1$NUS, \, ~$^2$CUHK, \, ~$^3$NTU, \, ~$^4$UR}
\\ 
{\vspace{0.4cm}
Survey Project: \url{https://github.com/yaotingwangofficial/Awesome-MCoT}
}
}

\begin{document}

\maketitle
 
\setcounter{footnote}{0}

\vspace{-5mm}
\begin{abstract}
By extending the advantage of chain-of-thought (CoT) reasoning in human-like step-by-step processes to multimodal contexts, multimodal CoT (MCoT) reasoning has recently garnered significant research attention, especially in the integration with multimodal large language models (MLLMs).  
Existing MCoT studies design various methodologies and innovative reasoning paradigms to address the unique challenges of image, video, speech, audio, 3D, and structured data across different modalities, achieving extensive success in applications such as robotics, healthcare, autonomous driving, and multimodal generation.  
However, MCoT still presents distinct challenges and opportunities that require further focus to ensure consistent thriving in this field, where unfortunately an up-to-date review of this domain is lacking.  
To bridge this gap, we present the first systematic survey of MCoT reasoning, elucidating the relevant foundational concepts and definitions.  
We offer a comprehensive taxonomy and an in-depth analysis of current methodologies from diverse perspectives across various application scenarios. 
Furthermore, we provide insights into existing challenges and future research directions, aiming to foster innovation toward multimodal AGI.
\end{abstract}
\vspace{-0.3cm}
\hspace{30pt}
{\keywords{Multimodal Reasoning, Chain-of-Thought, Multimodal Large Language Models}}

\begin{figure*}[!h]
  \centering
   \includegraphics[width=0.99\linewidth,height=0.55\linewidth]{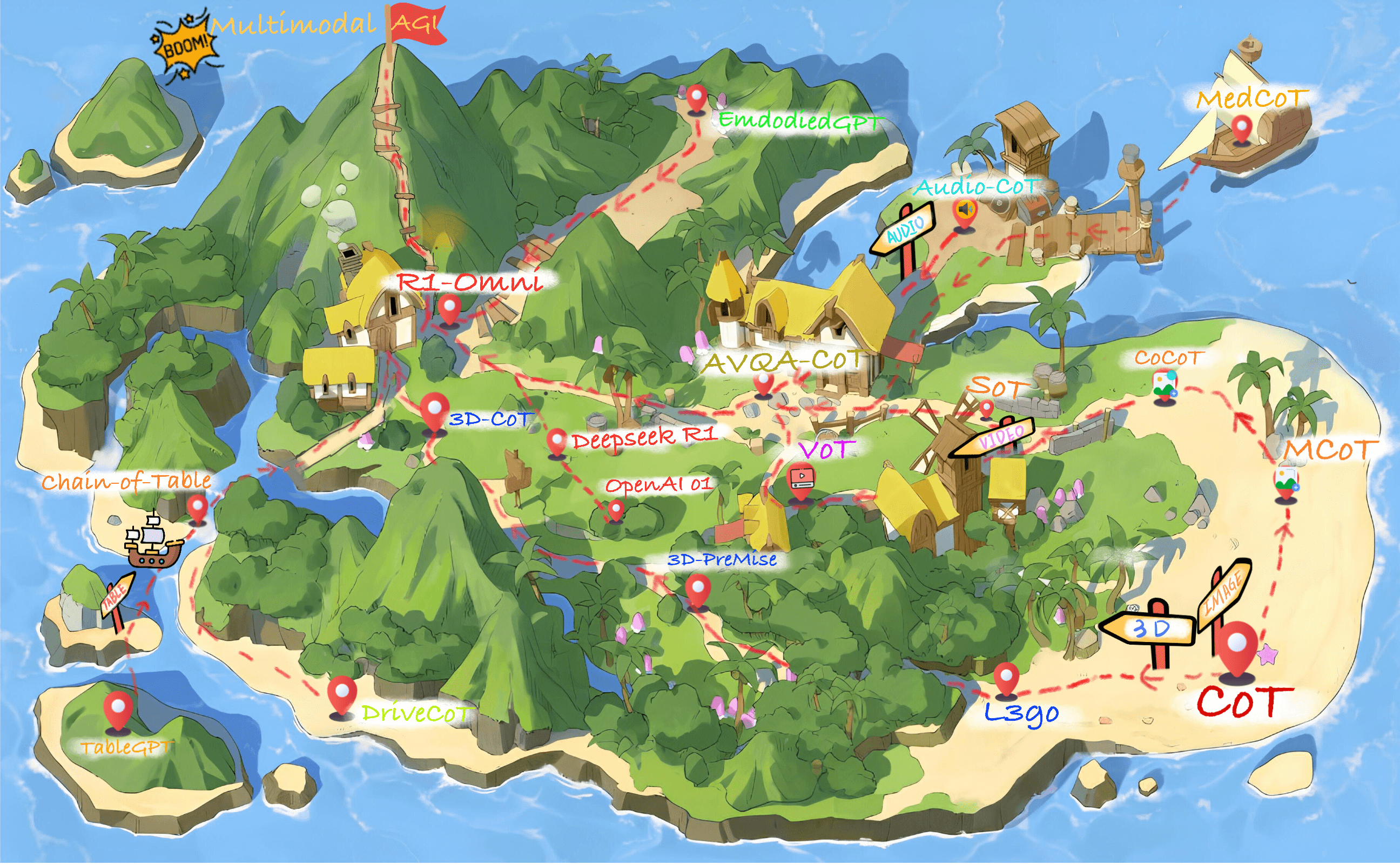}
   \vspace{-3mm}
   \label{fig:teaser}
\end{figure*}

\newpage
{
  \hypersetup{linktoc=page} 
  \tableofcontents
}

\newpage

\begin{CJK}{UTF8}{gbsn}

\begin{figure}[t!]
    \centering
    \includegraphics[width=1\textwidth]{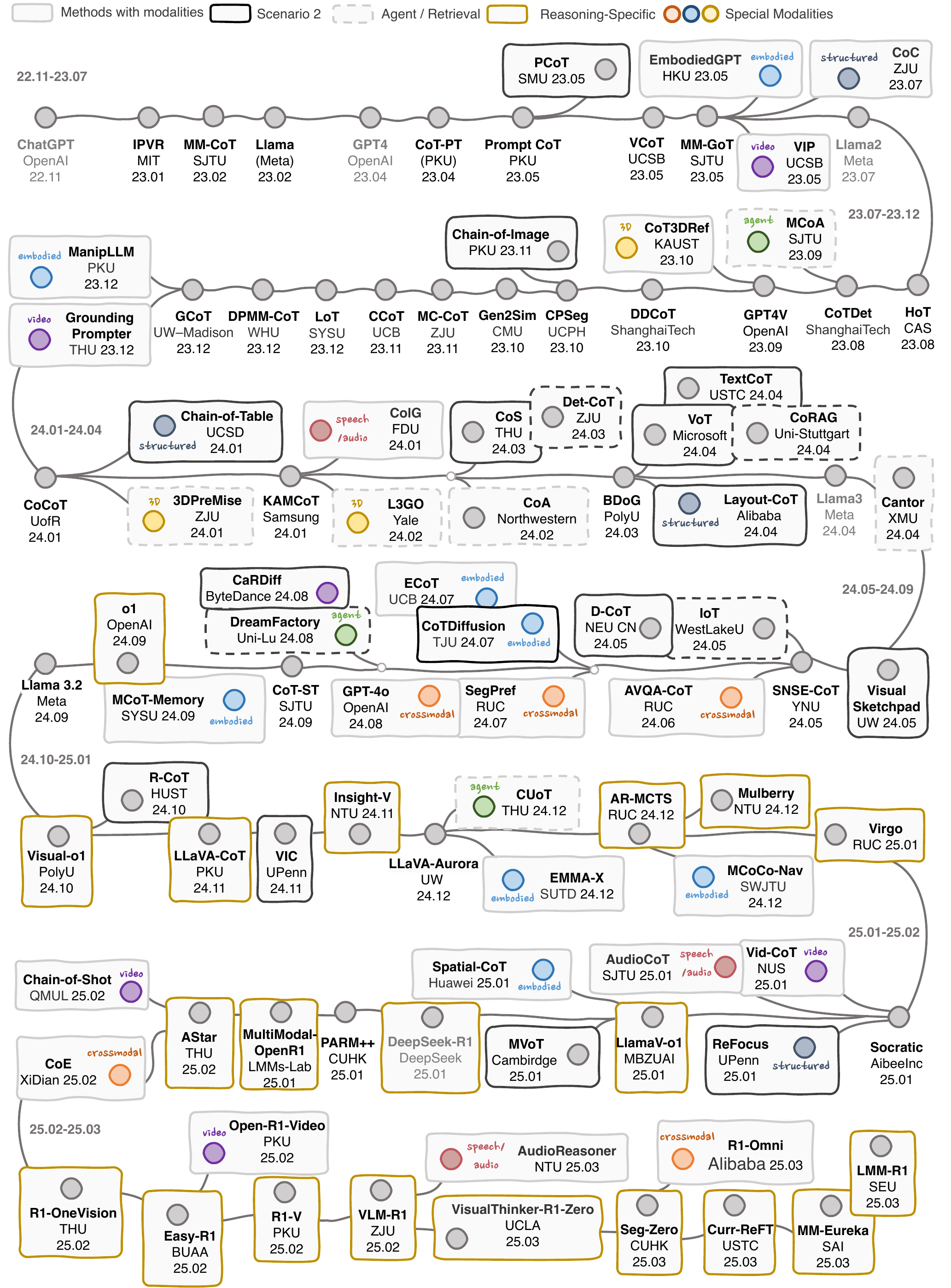}
    \caption{Developing timeline of Multimodal Chain-of-Thought (MCoT) reasoning. Models with names in \textcolor{gray}{gray} are text-only LLMs. For clarity, the models in the figure are assumed to include the image modality by default, unless specified with special modalities indicated by colored circles.}
    \vspace{-3mm}
    \label{fig:timeline}
\end{figure}

\vspace{-2mm}
\section{Introduction}
\vspace{-2mm}

The emergence of large language models (LLMs) \cite{Qwen2-abs-2407-10671,ChatGLM-abs-2406-12793,Llama-2-abs-2307-09288,Yi-abs-2403-04652,PaLM-2-abs-2305-10403,Phi-3-abs-2404-14219,team2023gemini} has ushered in an unprecedented era in the artificial intelligence (AI) field. 
It has been long recognized the necessity of aligning with the inherently multimodal nature of real-world environments, and correspondingly, the AI field evolves from LLMs to multimodal LLMs (MLLMs) \cite{LLaVA-UHD-GuoXYCNGCLH24,LLaVA-CoT-abs-2411-10440,LLaVA-Med-LiWZULYNPG23,Qwen2-VL-abs-2409-12191,Monkey-LiYLMZYSLB24,liu2023llava,NExT-GPT-Wu0Q0C24,Vitron-0001WZCY24,chu2024qwen2audio,maaz2023video,zhang2023videollama}, integrating diverse modalities into language intelligence.
Achieving human-level intelligence requires transcending basic perceptual capabilities to attain sophisticated cognitive reasoning—a hallmark of human cognition that enables iterative reasoning through contextual understanding and self-correction.
Inspired by this observation, in-context learning (ICL) techniques have empowered LLMs to demonstrate stepwise reasoning—commonly known as chain-of-thought (CoT) reasoning mechanisms \cite{wei2022CoT,zhang2022automatic,0002Z24,MadaanTGHGW0DPY23,besta2024GoT,YaoYZS00N23}. 
This technique enables models to break down problems into a series of intermediate steps, enhancing both transparency in decision-making and performance on intricate reasoning tasks.
The remarkable success of CoT reasoning on a wide range of downstream complex tasks has driven its widespread adoption across academia and industry.
Especially the recent advancements implicitly integrating this capability into cutting-edge systems like OpenAI's o1/o3 \cite{OpenAI-o1-abs-2412-16720} and DeepSeek R1 \cite{DeepSeek-R1-abs-2501-12948} has garnered widespread attention.

The integration of CoT reasoning into multimodal contexts has then catalyzed transformative progress in AI, giving rise to Multimodal Chain-of-Thought (MCoT) reasoning \cite{abs-2410-14668,Chen0ZC0C24}. 
The MCoT topic has generated a spectrum of innovative outcomes due to both the CoT attributes and the heterogeneous nature of cross-modal data interactions.
On one hand, the original CoT framework has evolved into advanced reasoning architectures incorporating hierarchical thought structures, from linear sequences \cite{wei2022CoT} to graph-based representations \cite{besta2024GoT}.
On the other hand, unlike the unimodal text setting, diverse modalities such as visual, auditory, and spatiotemporal data demand specialized processing strategies—visual reasoning requires precise perception and analysis of static scenes, object relationship, while video understanding necessitates robust temporal dynamics modeling. 
These requirements have spurred the development of array of sophisticated MCoT methodologies that adapt reasoning processes to modality-specific characteristics, such as  Multimodal-CoT \cite{zhang2023multimodal}, MVoT \cite{li2025imagine}, Video-of-Thought \cite{fei2024video}, Audio-CoT \cite{li2024avqa_cot}, Cot3DRef \cite{abdelrahman2023cot3dref} and PARM++ \cite{guo2025can}. 
The demonstrated effectiveness of MCoT has also led to its successful application in critical domains such as autonomous driving \cite{DriveVLM-abs-2402-12289,DiLu-WenF0C0CDS0024,abs-2405-01533,abs-2405-18361}, embodied AI \cite{mu2023embodiedgpt,LinLL24,zawalski2024robotic}, robotics \cite{SinghBMGXTFTG23,abs-2311-14379,RanaHGA0S23,ZitkovichYXXXXW23} and healthcare \cite{abs-2404-03264,GoyalRRYZCNW24,abs-2402-13408,JinCVHCK24,abs-2409-19487}, positioning it as a foundational technology for achieving multimodal AGI.

In recent years, research on MCoT has attracted growing attention. 
Figure \ref{fig:timeline} presents a comprehensive timeline of key milestones in this emerging field. 
Despite its promising potential in enhancing multimodal reasoning, MCoT also poses significant challenges and leaves several critical questions unanswered—for example, determining the most effective strategies for leveraging varied multimodal context, designing CoT processes that truly enhance MLLMs' reasoning capabilities, and implementing implicit reasoning within these models. 
Notably, the absence of comprehensive surveys hinders knowledge consolidation in this emerging field. 
To bridge this critical gap, this paper provides the first systematic overview of MCoT reasoning, providing a structured analysis of technological development, methodologies, practical applications, and future directions. 
We hope this survey will serve as an authoritative reference, spurring further innovation and progress in this rapidly evolving domain.

\vspace{-2mm}
\subsection{Contributions}

\begin{compactitem}
    \item \textbf{First Survey:} This paper represents the first survey dedicated to an inaugural thorough review of MCoT reasoning.
    
    \item \textbf{Comprehensive Taxonomy:} We propose a meticulous taxonomy (cf. \cref{fig:taxonomy}) that categorizes the diverse approaches in MCoT research.
    
    \item \textbf{Frontiers and Future Directions:} We discuss emerging challenges and outline promising avenues for future research.
    
    \item \textbf{Resource Sharing:} We compile and make publicly available all relevant resources to support and accelerate progress within the research community.
    
\end{compactitem}

\vspace{-2mm}
\subsection{Survey Organization}
\vspace{-2mm}

The remainder of this survey is organized as follows. 
We begin by introducing the fundamental concepts and background knowledge related to MCoT ($\S$\ref{Background and Preliminary}). 
We then review the state-of-the-art research in MCoT across different modalities ($\S$\ref{MCoT Reasoning Under Various Modality}). 
Next, we provide a taxonomy and consolidate the mainstream methods in MCoT under various perspectives ($\S$\ref{Methodologies in MCoT Reasoning}). 
Following this, we summarize the extensive downstream applications of MCoT ($\S$\ref{sec:applications}). 
Subsequently, we present an overview of datasets and benchmarks from multiple perspectives ($\S$\ref{MCoT Datasets and Benchmarks}). 
Finally, we discuss the challenges and future directions in this field ($\S$\ref{Limitations}).

\tikzstyle{my-box}=[
    rectangle,
    draw=gray,
    rounded corners,
    text opacity=1,
    minimum height=1.5em,
    minimum width=5em,
    inner sep=2pt,
    align=center,
    fill opacity=.5,
    line width=0.8pt,
]
\tikzstyle{leaf}=[my-box, minimum height=1.5em,
    fill=pink!10, text=black, align=left,font=\normalsize,
    inner xsep=2pt,
    inner ysep=4pt,
    line width=0.8pt,
]

\definecolor{c1}{RGB}{93,191,237} 
\definecolor{c2}{RGB}{237,110,106} 
\definecolor{c3}{RGB}{240,154,69} 
\definecolor{c4}{RGB}{108,222,157} 
\definecolor{c5}{RGB}{205,180,243} 
\definecolor{c6}{RGB}{97,218,184} 
\definecolor{c7}{RGB}{226,115,150} 
\definecolor{c8}{RGB}{201,116,201} 
\definecolor{c9}{RGB}{23,182,179} 
\definecolor{c10}{RGB}{242,157,108} 

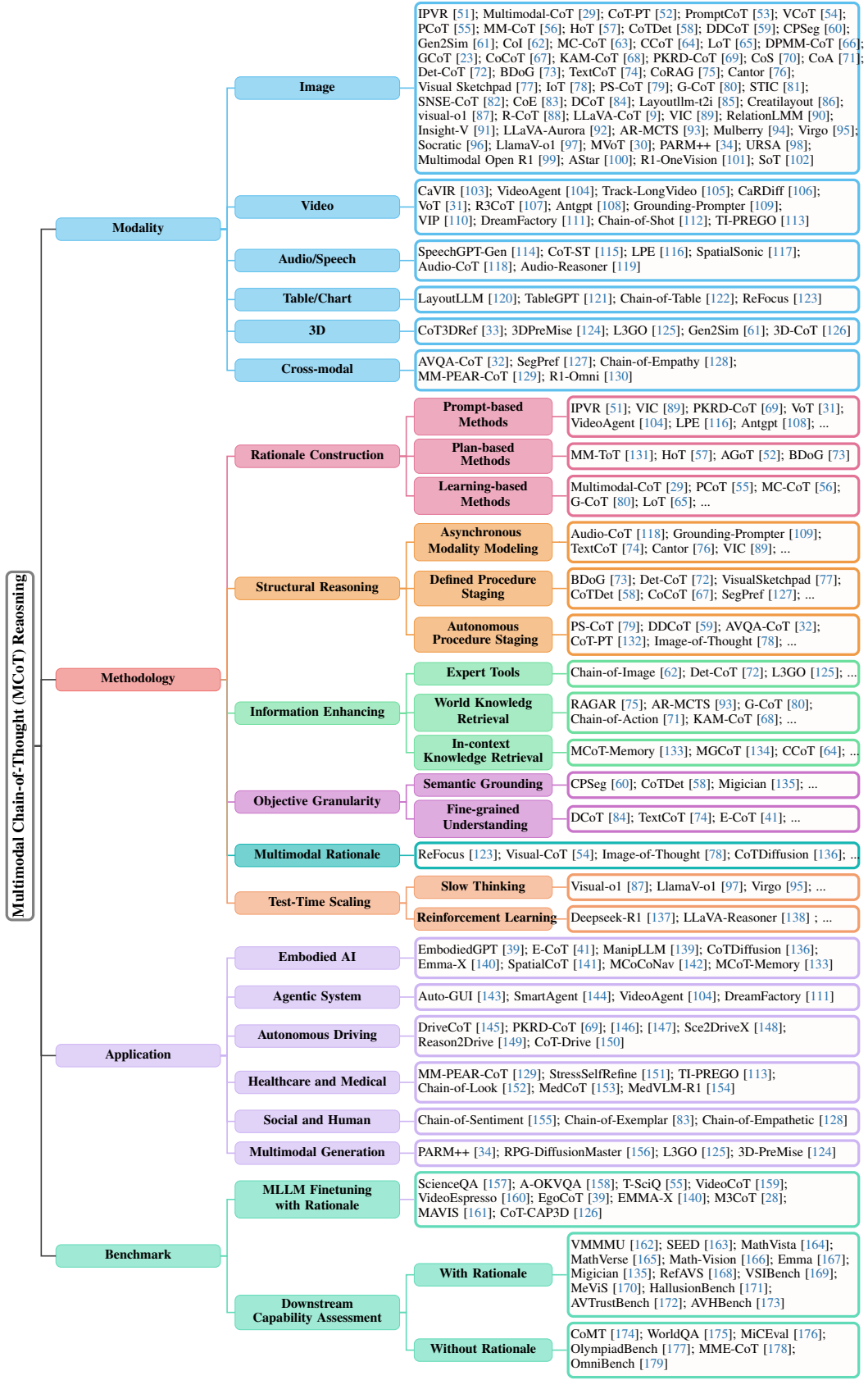
\begin{figure*}[!pt]
    \centering
    \resizebox{\textwidth}{!}{
        \begin{forest}
            forked edges,
            for tree={
                grow=east,
                reversed=true,
                anchor=base west,
                parent anchor=east,
                child anchor=west,
                base=center,
                font=\large,
                rectangle,
                draw=gray,
                rounded corners,
                align=left,
                text centered,
                minimum width=4em,
                edge+={darkgray, line width=1pt},
                s sep=3pt,
                inner xsep=2pt,
                inner ysep=3pt,
                line width=0.8pt,
                ver/.style={rotate=90, child anchor=north, parent anchor=south, anchor=center},
            },
            where level=1{text width=12em,font=\normalsize,}{},
            where level=2{text width=12em,font=\normalsize,}{},
            where level=3{text width=10em,font=\normalsize,}{},
            where level=4{text width=35em,font=\normalsize,}{},
            where level=5{text width=10em,font=\normalsize,}{},
            [
                \textbf{Multimodal Chain-of-Thought (MCoT) Reaosning}, ver, line width=0.7mm
                [
                    \textbf{Modality}, fill=c1!60, draw=c1, line width=0mm
                    [
                        \textbf{Image}, fill=c1!60, draw=c1, line width=0mm, edge={c1}
                        [
                            IPVR \cite{chen2023see}; Multimodal-CoT \cite{zhang2023multimodal}; CoT-PT \cite{yang2024AGoT}; PromptCoT \cite{yao2024promptcot}; VCoT \cite{rose2023visual}; \\
                            PCoT \cite{wang2024t}; MM-CoT \cite{tan2024boosting}; HoT \cite{yao2023HoT}; CoTDet \cite{tang2023cotdet}; DDCoT \cite{zheng2023ddcot}; CPSeg \cite{li2024cpseg};\\
                             Gen2Sim \cite{katara2024gen2sim}; CoI \cite{meng2023chain}; MC-CoT \cite{wei2024mccot}; CCoT \cite{mitra2024compositional}; LoT \cite{zhong2024let}; DPMM-CoT \cite{he2024multi};  \\
                            GCoT \cite{besta2024GoT}; CoCoT \cite{zhang2024cocot}; KAM-CoT \cite{mondal2024kam}; PKRD-CoT \cite{luo2024pkrd}; CoS \cite{liu2024chain}; CoA \cite{pan2024chain}; \\
                            Det-CoT \cite{wu2024dettoolchain}; BDoG \cite{zheng2024picture}; TextCoT \cite{luan2024textcot}; CoRAG \cite{khaliq2024ragar}; Cantor \cite{gao2024cantor}; \\
                             Visual Sketchpad \cite{hu2024visual}; IoT \cite{zhou2024image}; PS-CoT \cite{li2025ps}; G-CoT \cite{ma2024dolphins}; STIC \cite{deng2024stic}; \\
                             SNSE-CoT \cite{zheng2024enhancing}; CoE \cite{luo2024chain}; DCoT \cite{jia2024dcot}; Layoutllm-t2i \cite{qu2023layoutllm}; Creatilayout \cite{zhang2024creatilayout}; \\
                             visual-o1 \cite{ni2024visual_o1}; R-CoT \cite{deng2024r}; LLaVA-CoT \cite{LLaVA-CoT-abs-2411-10440}; VIC \cite{zheng2024thinking}; RelationLMM \cite{xie2025relationlmm}; \\
                             Insight-V \cite{dong2024insight}; LLaVA-Aurora \cite{bigverdi2024perception};
                            AR-MCTS \cite{dong2024progressive_retrieval}; Mulberry \cite{yao2024mulberry}; Virgo \cite{du2025virgo}; \\
                            Socratic \cite{hu2025socratic}; 
                            LlamaV-o1 \cite{thawakar2025llamav_o1}; MVoT \cite{li2025imagine};  PARM++ \cite{guo2025can}; URSA \cite{luo2025ursa}; \\
                            Multimodal Open R1 \cite{multimodal-open-r1}; AStar \cite{wu2025boosting}; R1-OneVision \cite{R1-OneVision}; SoT \cite{aytes2025sketchofthought},leaf, text width=33em, draw=c1, line width=0.7mm, edge={c1}
                        ]
                    ]
                    [
                        \textbf{Video}, fill=c1!60, draw=c1, line width=0mm, edge={c1}
                        [
                            CaVIR \cite{li2023intentqa}; VideoAgent \cite{wang2024videoagent}; Track-LongVideo \cite{sun2024hallucination};  CaRDiff \cite{tang2024cardiff}; \\
                            VoT \cite{fei2024video}; R3CoT \cite{hong2025following}; Antgpt \cite{zhao2023antgpt}; Grounding-Prompter \cite{chen2023grounding}; \\
                            VIP \cite{himakunthala2023let}; DreamFactory \cite{xie2024dreamfactory}; Chain-of-Shot \cite{hu2025cos};   TI-PREGO \cite{plini2024ti}, leaf, text width=33em, draw=c1, line width=0.7mm, edge={c1}
                        ]
                    ]
                    [
                        \textbf{Audio/Speech}, fill=c1!60, draw=c1, line width=0mm, edge={c1}
                        [
                            SpeechGPT-Gen \cite{zhang2024speechgpt}; CoT-ST \cite{du2024cot_st}; LPE \cite{xie2025leveraging}; SpatialSonic \cite{sun2024spatial_audio}; \\
                            Audio-CoT \cite{ma2025audio_cot}; Audio-Reasoner \cite{xie2025audio},leaf, text width=33em, draw=c1, line width=0.7mm, edge={c1}
                        ]
                    ]
                    [
                        \textbf{Table/Chart}, fill=c1!60, draw=c1, line width=0mm, edge={c1}
                        [
                            LayoutLLM \cite{luo2024layoutllm}; TableGPT \cite{zha2023tablegpt}; Chain-of-Table \cite{wang2024chain_of_table}; ReFocus \cite{fu2025refocus},leaf, text width=33em, draw=c1, line width=0.7mm, edge={c1}
                        ]
                    ]
                    [
                        \textbf{3D}, fill=c1!60, draw=c1, line width=0mm, edge={c1}
                        [
                            CoT3DRef \cite{abdelrahman2023cot3dref}; 3DPreMise \cite{yuan20243d_PreMise}; L3GO \cite{yamada2024Co3D_Thoughts}; Gen2Sim \cite{katara2024gen2sim}; 3D-CoT \cite{chen2025integrating},leaf, text width=33em, draw=c1, line width=0.7mm, edge={c1}
                        ]
                    ]
                    [
                        \textbf{Cross-modal}, fill=c1!60, draw=c1, line width=0mm, edge={c1}
                        [
                            AVQA-CoT \cite{li2024avqa_cot}; SegPref \cite{wang2024avs_cot};  Chain-of-Empathy \cite{zhang2025CoEmpathetic}; \\
                            MM-PEAR-CoT \cite{li2024multimodal_emotion}; R1-Omni \cite{zhao2025r1omni},leaf, text width=33em, draw=c1, line width=0.7mm, edge={c1}
                        ]
                    ]
                ]
                [
                    \textbf{Methodology},
                    fill=c2!60, draw=c2, line width=0mm
                    [
                        \textbf{Rationale Construction}, align=center, fill=c7!60, draw=c7, line width=0mm, edge={c7}
                        [
                            \shortstack{\textbf{Prompt-based} \\ \textbf{Methods}}, align=center, fill=c7!60, draw=c7, line width=0mm, edge={c7}
                            [
                                IPVR \cite{chen2023see}; VIC \cite{zheng2024thinking}; PKRD-CoT \cite{luo2024pkrd}; VoT \cite{fei2024video}; \\
                                VideoAgent \cite{wang2024videoagent}; LPE \cite{xie2025leveraging}; Antgpt \cite{zhao2023antgpt}; ... ,leaf, text width=21.5em, draw=c7, line width=0.7mm, edge={c7}
                            ]
                        ]
                        [
                            \shortstack{\textbf{Plan-based} \\ \textbf{ Methods}}, align=center, fill=c7!60, draw=c7, line width=0mm, edge={c7}
                            [
                                MM-ToT \cite{gomez2023mmtot}; HoT \cite{yao2023HoT}; AGoT \cite{yang2024AGoT}; BDoG \cite{zheng2024picture}
                                ,leaf, text width=21.5em, draw=c7, line width=0.7mm, edge={c7}
                            ]
                        ]
                        [
                            \shortstack{\textbf{Learning-based} \\ \textbf{ Methods}}, align=center, fill=c7!60, draw=c7, line width=0mm, edge={c7}
                            [
                                Multimodal-CoT \cite{zhang2023multimodal}; PCoT \cite{wang2024t}; MC-CoT \cite{tan2024boosting}; \\
                                G-CoT \cite{ma2024dolphins}; LoT \cite{zhong2024let}; ...
                                ,leaf, text width=21.5em, draw=c7, line width=0.7mm, edge={c7}
                            ]
                        ]
                    ]
                    [
                        \textbf{Structural Reasoning}, fill=c3!60, draw=c3, line width=0mm, edge={c3}
                        [
                            \shortstack{\textbf{Asynchronous } \\ \textbf{Modality Modeling}}, fill=c3!60, draw=c3, line width=0mm, edge={c3}
                            [
                                Audio-CoT \cite{ma2025audio_cot}; Grounding-Prompter \cite{chen2023grounding}; \\
                                TextCoT \cite{luan2024textcot}; Cantor \cite{gao2024cantor}; VIC \cite{zheng2024thinking}; ...
                            ,leaf, text width=21.5em, draw=c3, line width=0.7mm, edge={c3}
                            ]
                        ]
                        [
                            \shortstack{\textbf{Defined Procedure} \\  \textbf{Staging}}, fill=c3!60, draw=c3, line width=0mm, edge={c3}
                            [
                                BDoG \cite{zheng2024picture}; Det-CoT \cite{wu2024dettoolchain}; VisualSketchpad \cite{hu2024visual}; \\
                                CoTDet \cite{tang2023cotdet}; CoCoT \cite{zhang2024cocot}; SegPref \cite{wang2024avs_cot}; ... 
                            ,leaf, text width=21.5em, draw=c3, line width=0.7mm, edge={c3}
                            ]
                        ]
                        [
                            \shortstack{\textbf{Autonomous} \\ \textbf{ Procedure Staging}}, fill=c3!60, draw=c3, line width=0mm, edge={c3}
                            [
                                PS-CoT \cite{li2025ps}; DDCoT \cite{zheng2023ddcot}; AVQA-CoT \cite{li2024avqa_cot}; \\
                                CoT-PT \cite{ge2023chain}; Image-of-Thought \cite{zhou2024image}; ... 
                            ,leaf, text width=21.5em, draw=c3, line width=0.7mm, edge={c3}
                            ]
                        ]
                    ]
                    [
                        \textbf{Information Enhancing}, fill=c4!60, draw=c4, line width=0mm, edge={c4}
                        [
                            \textbf{Expert Tools }, fill=c4!60, draw=c4, line width=0mm, edge={c4}
                            [
                                Chain-of-Image \cite{meng2023chain}; 
                                Det-CoT \cite{wu2024dettoolchain};  L3GO \cite{yamada2024Co3D_Thoughts}; ...
                                , leaf, text width=21.5em, draw=c4, line width=0.7mm, edge={c4}
                            ]
                        ]
                        [
                            \shortstack{\textbf{World Knowledg} \\ \textbf{Retrieval }}, fill=c4!60, draw=c4, line width=0mm, edge={c4}
                            [
                                RAGAR \cite{khaliq2024ragar}; AR-MCTS \cite{dong2024progressive_retrieval};
                                G-CoT \cite{ma2024dolphins}; \\
                                Chain-of-Action \cite{pan2024chain}; KAM-CoT \cite{mondal2024kam}; ...
                                , leaf, text width=21.5em, draw=c4, line width=0.7mm, edge={c4}
                            ]
                        ]
                        [
                            \shortstack{\textbf{In-context } \\ \textbf{Knowledge Retrieval}}, fill=c4!60, draw=c4, line width=0mm, edge={c4}
                            [
                                MCoT-Memory \cite{liang2024memory_driven}; MGCoT \cite{yao2023beyond}; CCoT \cite{mitra2024compositional}; ...
                                , leaf, text width=21.5em, draw=c4, line width=0.7mm, edge={c4}
                            ]
                        ]
                    ]
                    [
                        \textbf{Objective Granularity}, fill=c8!60, draw=c8, line width=0mm, edge={c8}
                        [
                            \textbf{Semantic Grounding}, fill=c8!60, draw=c8, line width=0mm, edge={c8}
                            [
                                CPSeg \cite{li2024cpseg}; CoTDet \cite{tang2023cotdet}; Migician \cite{li2025migician}; ...
                                ,leaf, text width=21.5em, draw=c8, line width=0.7mm, edge={c8}
                            ]
                        ]
                        [
                            \shortstack{\textbf{Fine-grained} \\ \textbf{ Understanding}}, fill=c8!60, draw=c8, line width=0mm, edge={c8}
                            [
                                DCoT \cite{jia2024dcot}; TextCoT \cite{luan2024textcot}; E-CoT \cite{zawalski2024robotic}; ...
                                ,leaf, text width=21.5em, draw=c8, line width=0.7mm, edge={c8}
                            ]
                        ]
                    ]
                    [
                        \textbf{Multimodal  Rationale}, fill=c9!60, draw=c9, line width=0mm, edge={c9}
                        [
                            ReFocus \cite{fu2025refocus}; Visual-CoT \cite{rose2023visual}; Image-of-Thought \cite{zhou2024image}; CoTDiffusion \cite{ni2024generate}; ... ,leaf, text width=33em, draw=c9, line width=0.7mm, edge={c9}
                        ]
                    ]
                    [
                        \textbf{ Test-Time Scaling}, fill=c10!60, draw=c10, line width=0mm, edge={c10}
                        [
                            \textbf{Slow Thinking}, fill=c10!60, draw=c10, line width=0mm, edge={c10}
                            [
                                Visual-o1 \cite{ni2024visual_o1}; LlamaV-o1 \cite{thawakar2025llamav_o1}; Virgo \cite{du2025virgo}; ...
                                ,leaf, text width=21.5em, draw=c10, line width=0.7mm, edge={c10}
                            ]
                        ]
                        [
                            \textbf{Reinforcement Learning}, fill=c10!60, draw=c10, line width=0mm, edge={c10}
                            [
                                Deepseek-R1 \cite{guo2025deepseekr1}; LLaVA-Reasoner \cite{zhang2024llavareasoner} ; ...
                                ,leaf, text width=21.5em, draw=c10, line width=0.7mm, edge={c10}
                            ]
                        ]
                    ]
                ] 
                [
                    \textbf{Application}, fill=c5!60, draw=c5, line width=0mm
                    [
                        \textbf{Embodied AI}, fill=c5!60, draw=c5, line width=0mm, edge={c5}
                        [
                        EmbodiedGPT \cite{mu2023embodiedgpt};
                        E-CoT \cite{zawalski2024robotic}; ManipLLM \cite{li2024manipllm}; CoTDiffusion \cite{ni2024generate}; \\
                        Emma-X \cite{sun2024emmax};
                        SpatialCoT \cite{liu2025spatialcot}; MCoCoNav \cite{shen2024enhancing}; MCoT-Memory \cite{liang2024memory_driven}, leaf, text width=33em, draw=c5, line width=0.7mm, edge={c5}
                        ]
                    ]
                    [
                        \textbf{Agentic System}, fill=c5!60, draw=c5, line width=0mm, edge={c5}
                        [
                        Auto-GUI \cite{zhang2023CoAction}; SmartAgent \cite{zhang2024CoUser}; VideoAgent \cite{wang2024videoagent}; DreamFactory \cite{xie2024dreamfactory}, leaf, text width=33em, draw=c5, line width=0.7mm, edge={c5}
                        ]
                    ]
                    [
                        \textbf{Autonomous Driving}, fill=c5!60, draw=c5, line width=0mm, edge={c5}
                        [
                        DriveCoT \cite{wang2024drivecot}; PKRD-CoT \cite{luo2024pkrd}; \cite{cui2024receive}; \citep{ma2024learning}; Sce2DriveX \cite{zhao2025sce2drivex}; \\
                        Reason2Drive \cite{nie2024reason2drive}; CoT-Drive \cite{liao2025cotdrive}
                        , leaf, text width=33em, draw=c5, line width=0.7mm, edge={c5}
                        ]
                    ]
                    [
                        \textbf{Healthcare and Medical}, fill=c5!60, draw=c5, line width=0mm, edge={c5}
                        [
                         MM-PEAR-CoT \cite{li2024multimodal_emotion}; StressSelfRefine \cite{dai2024interpretable};  TI-PREGO \cite{plini2024ti};\\
                         Chain-of-Look \cite{xi2023chainoflook}; MedCoT \cite{liu2024medcot}; MedVLM-R1 \cite{pan2025medvlm}
                        , leaf, text width=33em, draw=c5, line width=0.7mm, edge={c5}
                        ]
                    ]
                    [
                        \textbf{Social and Human}, fill=c5!60, draw=c5, line width=0mm, edge={c5}
                        [
                         Chain-of-Sentiment \cite{luo2024panosent}; Chain-of-Exemplar \cite{luo2024chain}; Chain-of-Empathetic \cite{zhang2025CoEmpathetic}
                        , leaf, text width=33em, draw=c5, line width=0.7mm, edge={c5}
                        ]
                    ]
                    [
                        \textbf{Multimodal Generation}, fill=c5!60, draw=c5, line width=0mm, edge={c5}
                        [
                        PARM++ \cite{guo2025can}; RPG-DiffusionMaster \cite{yang2024mastering}; L3GO \cite{yamada2024Co3D_Thoughts}; 3D-PreMise \cite{yuan20243d_PreMise}
                        , leaf, text width=33em, draw=c5, line width=0.7mm, edge={c5}
                        ]
                    ]
                ]
                [
                    \textbf{Benchmark}, fill=c6!60, draw=c6, line width=0mm
                    [
                        \shortstack{\textbf{MLLM Finetuning} \\ \textbf{with Rationale}},  fill=c6!60, draw=c6, line width=0mm, edge={c6}
                        [
                        ScienceQA \cite{lu2022scienceqa}; A-OKVQA \cite{schwenk2022okvqa}; T-SciQ \cite{wang2024t}; VideoCoT \cite{wang2024videocot}; \\
                        VideoEspresso \cite{han2024videoespresso}; EgoCoT \cite{mu2023embodiedgpt};  EMMA-X \cite{sun2024emmax}; M3CoT \cite{Chen0ZC0C24}; \\
                        MAVIS \cite{zhang2024mavis}; CoT-CAP3D \cite{chen2025integrating}
                        , leaf, text width=33em, draw=c6, line width=0.7mm, edge={c5}
                        ]
                    ]
                    [
                        \shortstack{\textbf{Downstream} \\ \textbf{Capability Assessment}}, fill=c6!60, draw=c6, line width=0mm, edge={c6}
                        [
                            \textbf{With Rationale}, fill=c6!60, draw=c6, line width=0mm, edge={c6}
                            [
                                VMMMU \cite{yue2024mmmu}; SEED \cite{li2024seed}; MathVista \cite{lu2023mathvista}; \\
                                MathVerse \cite{zhang2024mathverse};
                                Math-Vision \cite{wang2025mathvision}; Emma \cite{hao2025emma}; \\
                                Migician \cite{li2025migician};  RefAVS \cite{wang2024refavs}; 
                                VSIBench \cite{yang2024vsibench}; \\
                                MeViS \cite{ding2023mevis}; HallusionBench \cite{guan2024hallusionbench}; \\
                                AVTrustBench \cite{chowdhury2025avtrustbench}; 
                                AVHBench \cite{sung2024avhbench}
                                , leaf, text width=21.5em, draw=c6, line width=0.7mm, edge={c6}
                            ]
                        ]
                        [
                            \textbf{Without Rationale}, fill=c6!60, draw=c6, line width=0mm, edge={c6}
                            [
                                CoMT \cite{cheng2024comt};  WorldQA \cite{zhang2024worldqa};  MiCEval \cite{zhou2024miceval};  \\
                                OlympiadBench \cite{he2024olympiadbench};  MME-CoT \cite{jiang2025mmecot}; \\
                                OmniBench \cite{li2024omnibench}
                                , leaf, text width=21.5em, draw=c6, line width=0.7mm, edge={c6}
                            ]
                        ]
                    ]
                ]
            ]
        \end{forest}
    }
    \caption{Taxonomy of MCoT reasoning.}
    \label{fig:taxonomy}
\end{figure*}

\section{Background and Preliminary}
\label{Background and Preliminary}

Recent advancements in the scale of model pretraining have driven a significant shift in the application paradigm of language models, transitioning from the conventional ``\emph{pretrain-then-finetune}'' approach to a more adaptive ``\emph{pretrain-then-prompt}'' framework \cite{brown2020language, achiam2023gpt4, touvron2023llama1, touvron2023llama2, zhao2023survey}. 
Within this evolving landscape, researchers have explored innovative techniques to enhance the reasoning capabilities of LLMs for complex tasks, notably ICL \cite{brown2020language,wei2022emergent,qin2024factors} and CoT\footnote{For clarity and consistency, we use ``CoT'' to denote multi-step reasoning technologies and ``topology'' to describe distinct thought structures (\eg, chain or graph topologies). Specific approaches are assigned unique identifiers, such as vanilla CoT.} reasoning \cite{wei2022CoT}.
The essence of ICL lies in supplying task-relevant examples or demonstrations within the prompt, enabling LLMs to better interpret user intent and generate outputs aligned with expectations. 
This method leverages contextual guidance to steer the model toward task-appropriate responses. 
In contrast, CoT reasoning emulates human problem-solving by decomposing complex tasks into a sequence of manageable sub-tasks, systematically constructing solutions. 
The intermediate reasoning steps or trajectories, termed the rationale, elucidate the logical progression underlying the model's conclusions.
Building on this foundation, MCoT reasoning extends the CoT paradigm by incorporating diverse data modalities, such as images, videos, and audio. 
This augmentation broadens the scope of multi-step reasoning, enhancing its applicability to increasingly intricate scenarios.

\begin{table}[]
\fontsize{9.5}{11}\selectfont
\centering
\setlength{\tabcolsep}{2.5mm}
\begin{tabular}{lcp{8.5cm}}
\toprule
\rowcolor{headcolor} \bf Terms & \multicolumn{1}{c}{\bf Abbrev.} & \multicolumn{1}{c}{\bf Description} \\ \midrule
In-context Learning & ICL & Prompting LLMs with task-specific examples without additional explicit training. \\
\rowcolor{tablegrey} 
Chain-of-Thought & CoT & Prompting LLMs to reason step-by-step or breaks complex problems into logical steps. \\
Multimodal CoT & MCoT & Extends CoT to reason with multimodalities, \eg, audio, image. \\
\rowcolor{tablegrey} 
Cross-modal CoT &  & Reasoning with two or more multimodalities, \eg, audio-visual. \\
Thought &  & A single reasoning step in CoT. \\
\rowcolor{tablegrey} 
Rationale &  & Built upon multiple thoughts to support the final answer.\\
\bottomrule
\end{tabular}%
\vspace{1mm}
\caption{Interpretation of MCoT-related terms.}
\label{tab:terms}
\end{table}

\subsection{From CoT to MCoT}
We provide explanations of the terms related to MCoT in \cref{tab:terms}.
To formalize the MCoT framework, we begin by defining \(\mathcal{P}\), \(\mathcal{S}\), \(\mathcal{Q}\), \(\mathcal{A}\) and \(\mathcal{R}\) to represent the prompt, instruction, query, answer, and rationale, respectively. 
Each of these elements is represented as a sequence of language tokens, with length denoted as \(|\cdot|\). 
We also use lowercase letter to denote individual tokens, \eg, \( a_i \) refers to the \(i\)-th token of the answer \(\mathcal{A}\).
Next, we define a standard ICL process that integrates few-shot demonstration pairs, which can be expressed as follows:
\begin{equation}
\label{eq:icl_P}
\mathcal{P}_{ICL} = \{ \mathcal{S}, (x_1, y_1), \ldots, (x_n, y_n) \},
\end{equation}
where $\mathcal{P}_{ICL}$ represents the prompt for ICL, consisting of an instruction $\mathcal{S}$ along with $n$ demonstration pairs of questions $x$ and their corresponding answers $y$. 
Then, the probability of generating an answer sequence $\mathcal{A}$ given the prompt $\mathcal{P}_{ICL}$ and a query $\mathcal{Q}$ is mathematically defined as:
\begin{equation}
\label{eq:icl_A}
p(\mathcal{A} \mid \mathcal{P}_{ICL}, \mathcal{Q}) = \prod_{i=1}^{|\mathcal{A}|} \mathcal{F}(a_i \mid \mathcal{P}_{ICL}, \mathcal{Q}, a_{<i}),
\end{equation}
where $\mathcal{F}$ denotes the probabilistic language model. 
Note that when $n=0$, the process simplifies to the standard zero-shot prompting scenario.

Then, we can define the vanilla CoT as:
\begin{equation}
\label{eq:cot_P}
\mathcal{P}_{CoT} = \{ \mathcal{S}, (x_1, e_1, y_1), \ldots, (x_n, e_n, y_n) \},
\end{equation}
where \(\mathcal{P}_{CoT}\) denotes the prompt used for CoT reasoning, and \(e_i\) represents the example rational.
Next, we define the joint probability of generating an answer \(\mathcal{A}\) and rationale \(\mathcal{R}\) given the input prompt \(\mathcal{P}_{CoT}\) and a query \(\mathcal{Q}\):
\begin{equation}
\label{eq:cot_A_R}
p(\mathcal{A}, \mathcal{R} \mid \mathcal{P}_{CoT}, \mathcal{Q}) = p(\mathcal{A} \mid \mathcal{P}_{CoT}, \mathcal{Q}, \mathcal{R}) \cdot p(\mathcal{R} \mid \mathcal{P}_{CoT}, \mathcal{Q}),
\end{equation}
where the right side represents two conditional probabilities of generating the answer \(\mathcal{A}\) and rationale \(\mathcal{R}\), which can be defined as:
\begin{equation}
\label{eq:cot_R}
p(\mathcal{R} \mid \mathcal{P}_{CoT}, \mathcal{Q}) = \prod_{i=1}^{|\mathcal{R}|} \mathcal{F}(r_i \mid \mathcal{P}_{CoT}, \mathcal{Q}, r_{<i}),
\end{equation}
\begin{equation}
\label{eq:cot_A}
p(\mathcal{A} \mid \mathcal{P}_{CoT}, \mathcal{Q}, \mathcal{R}) = \prod_{i=1}^{|\mathcal{A}|} \mathcal{F}(a_i \mid \mathcal{P}_{CoT}, \mathcal{Q}, a_{<i}).
\end{equation}
In contrast to the ICL approach, as shown in Equation \eqref{eq:icl_A}, the CoT framework necessitates the generation of a rationale $\mathcal{R}$ prior to arriving at the answer $\mathcal{A}$, as reflected in Equations \eqref{eq:cot_R} and \eqref{eq:cot_A}.

When considering MCoT, it is crucial to highlight that, unlike CoT, MCoT incorporates multimodal information into the components \(\mathcal{P}\), \(\mathcal{Q}\), \(\mathcal{A}\), and \(\mathcal{R}\). 
However, it is not necessary for all these components to simultaneously encompass multimodal information. 
That is, given language-based input \( \mathcal{T} \) and language-excluded multimodal context \(\mathcal{M}\), we have \(\exists \vartheta \in \{\mathcal{P, Q, A, R}\}:\mathcal{M}(\vartheta)\). Hence we categorize MCoT into two distinct scenarios based on the composition of rationale: one relying exclusively on language-based information and another incorporating multimodal signals beyond linguistic content:

\begin{tcolorbox}[colback=blue!10, colframe=white]
\(\blacktriangleright\) \textbf{\emph{Scenario-1:}} MCoT with text-only thought to tackle multimodal input and output: \phantom{1111}\\
\begin{equation}\notag
    \mathcal{R} \in \mathcal{L} \,.
\end{equation}

\vspace{3mm}
\(\blacktriangleright\) \textbf{\emph{Scenario-2}}: MCoT with multimodal thought to tackle unimodal or multimodal scenes: \phantom{1111}
\begin{equation}\notag
    \mathcal{R} \in \{\mathcal{M}, \mathcal{M\oplus L}\} \,.
\end{equation}
\end{tcolorbox}
Scenario-1 aims to address tasks that involve multimodal information in either the input or output while utilizing a rationale composed solely of language. 
In contrast, Scenario-2 highlights the integration of given, retrieved or generated multimodal information within the rationale itself.

\subsection{Thought Paradigm}
\begin{figure}[h!]
    \centering
    \includegraphics[width=1\textwidth]{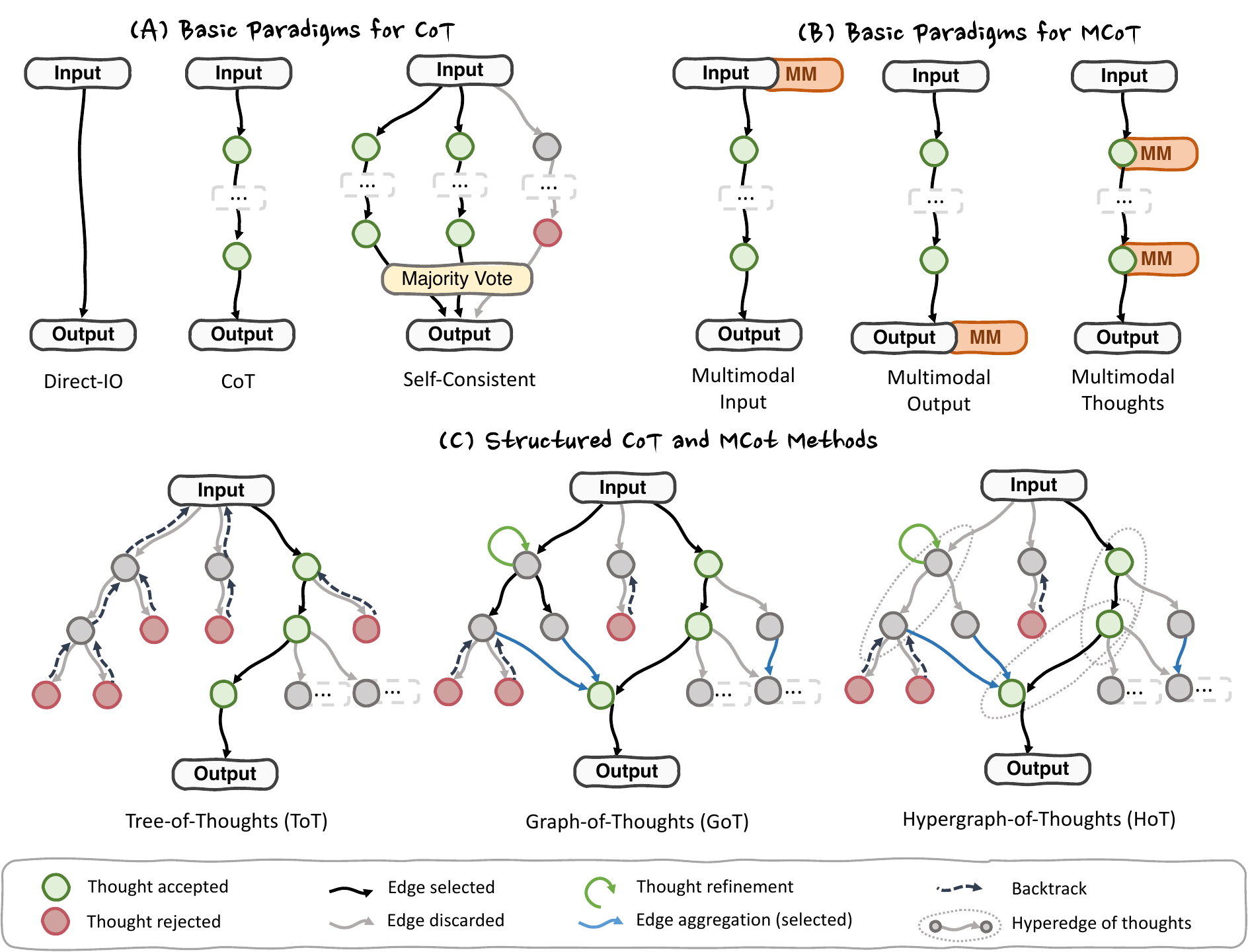}
    \caption{Different thought paradigms of CoT and MCoT.  
    }
    \label{fig:mcot_paradigm}
\end{figure}

Since the introduction of vanilla CoT \cite{wei2022CoT}, various paradigms have emerged to enhance multimodal and multi-step reasoning.
Based on the construction of thought generation during reasoning, the community categorizes the reasoning structures \cite{chu2023survey} or topologies \cite{besta2024topologies} into chain, tree, and graph types, as depicted in \cref{fig:mcot_paradigm}. 
In these topologies, thoughts are represented as nodes, with edges indicating dependencies between them \cite{besta2024topologies}.
Chain topologies \cite{wei2022CoT, kojima2022step_by_step, chen2022program} facilitate linear and sequential thought generation, progressively converging toward the final answer. 
However, chain topologies lack the capacity for in-depth exploration of individual thoughts during reasoning.

In contrast, tree topologies \cite{yao2023ToT, long2023ToT} enable exploration and backtracking within the reasoning process. 
At each node (\ie, thought) in a tree topology, a thought generator produces multiple child nodes, as illustrated on the left side of \cref{fig:mcot_paradigm}.C. 
These child nodes are then evaluated by a state evaluator, which assigns scores to them. 
These scores can be derived from the LLM itself or based on specific rules.
Search algorithms, such as breadth-first search (BFS) or depth-first search (DFS), then guide the tree’s expansion.

Graph topologies \cite{besta2024GoT} also allow for the generation of multiple child nodes from a single parent. However, they introduce cycles and N-to-1 connections, meaning that a single node can have multiple parent nodes. This facilitates aggregation among multiple nodes, as illustrated by the blue arrows in \cref{fig:mcot_paradigm}.
Hypergraph topologies \cite{yao2023HoT} extend graph topologies by employing hyperedges, which connect more than two thoughts. This structure inherently supports joint reasoning by integrating information from diverse modalities.
Furthermore, self-consistency \cite{wang2022cot_sc} can be seamlessly incorporated into various reasoning methods. 
For instance, using the chain topology as a baseline (\cref{fig:mcot_paradigm}.A), multiple chain-based reasoning processes can be executed in parallel, with the final answer determined by majority voting to ensure consistency across several rationales.
Overall, the evolution of reasoning topologies reflects a progression from linear dependencies to branching exploration, aggregation with refinement, and higher-order associations. 

\subsection{Multimodal LLMs}

\begin{figure}[h!]
    \centering
    \includegraphics[width=1\textwidth]{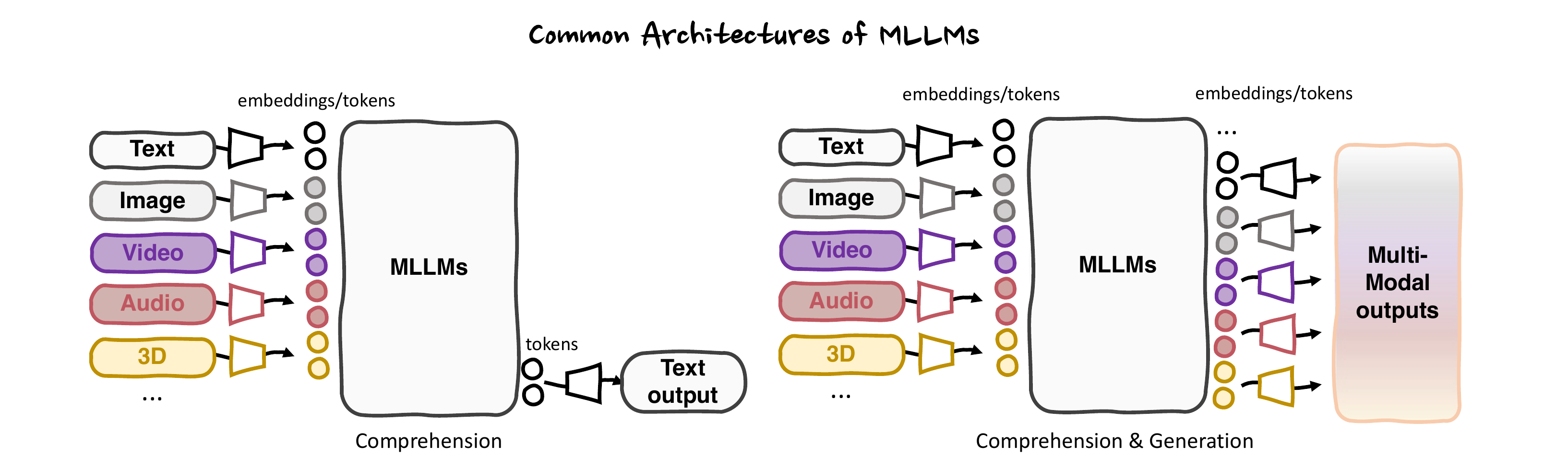}
    \caption{Common architectures for comprehension-only and comprehension-generation MLLMs.
    }
    \label{fig:mllm_tech}
\end{figure}

The release of models such as GPT-4V \cite{achiam2023gpt4}, Gemini 2.0 \cite{team2023gemini}, and Claude3 \cite{Anthropic2024claude3} has demonstrated remarkable capabilities in multimodal understanding, sparking significant interest in MLLMs within the research community.
Initial investigations into MLLMs focused on developing robust language models capable of interpreting multimodal content and generating textual responses. 
In the domain of image-text understanding, notable progress has been achieved with Visual Large Language Models (VLLMs) such as BLIP2 \cite{li2023blip}, OpenFlamingo \cite{awadalla2023openflamingo}, MiniGPT-4 \cite{zhu2023minigpt}, and LLaVA \cite{liu2023llava}. 
Concurrently, advancements in video-text understanding have emerged, with significant contributions from VideoChat \cite{li2023videochat} and Video-ChatGPT \cite{maaz2023video}. 
Audio and speech comprehension have also garnered attention, exemplified by models like Qwen-Audio \cite{chu2023qwen1audio,chu2024qwen2audio} and LLaSM \cite{shu2023llasm}. 
A noteworthy development is VideoLLaMA \cite{zhang2023videollama}, which leverages Qformer \cite{li2023blip} to enable both audio and video understanding.
In simple terms, mainstream MLLMs typically follow a consistent model architecture by processing multimodal embeddings or tokens into the decoder structure and generating contextually relevant outputs in an autoregressive manner, as shown in the left of \cref{fig:mllm_tech}.

Parallel to these works about multimodal understanding, research also explored multimodal content generation. 
In image generation, models such as Kosmos-2 \cite{peng2023kosmos2}, GILL \cite{koh2023gill}, Emu \cite{sun2023emu}, and MiniGPT-5 \cite{zheng2023minigpt5} have achieved breakthroughs. Audio generation has seen advancements with SpeechGPT \cite{zhang2023speechgpt,zhang2024speechgptgen} and AudioPaLM \cite{rubenstein2023audiopalm}, while video generation research, including CogVideo \cite{hong2022cogvideo}, VideoPoet \cite{kondratyuk2023videopoet}, Video-Lavit \cite{jin2024videolavit}, and StreamingT2V \cite{henschel2024streamingt2v}, has laid the groundwork for multimodal content creation.
The recent introduction of GPT-4o \cite{hurst2024gpt4o}, capable of both understanding and generating images and audio, has shifted attention toward ``any-to-any'' paradigm models. 
Prior works, such as NExT-GPT \cite{NExT-GPT-Wu0Q0C24} advances this objective for the first time by integrating multimodal adapters with various diffusion models. 
AnyGPT \cite{zhan2024anygpt} utilizes multimodal discrete tokens to facilitate the generation of diverse multimodal content. 
Subsequently, Mini-Omni2 \cite{xie2024miniomni1,xie2024miniomni2} introduces a command-based interruption mechanism, enhancing user interaction and aligning further with GPT-4o’s capabilities.
Compared to MLLMs that only support comprehension, as shown in \cref{fig:mllm_tech}, MLLMs that integrate both comprehension and generation either utilize an autoregressive approach to generate multimodal tokens \cite{zhan2024anygpt}, or connect decoders of varying modalities to decode multimodal embeddings \cite{NExT-GPT-Wu0Q0C24}.

Most recently, the release of the reasoning-focused OpenAI o1 \cite{jaech2024o1} model has drawn interest in enhancing reasoning capabilities through deliberate, extended processing and test-time scaling. 
Models such as Mulberry \cite{yao2024mulberry}, AStar \cite{wu2025boosting}, and LlamaV-o1 \cite{thawakar2025llamav_o1}, by adopting long-MCoT reasoning strategies, have demonstrated robust performance in multimodal reasoning, further advancing the field of multimodal understanding.

\section{MCoT Reasoning Under Various Modalities}
\label{MCoT Reasoning Under Various Modality}

\begin{figure}[!t]
    \centering
    \includegraphics[width=1\textwidth]{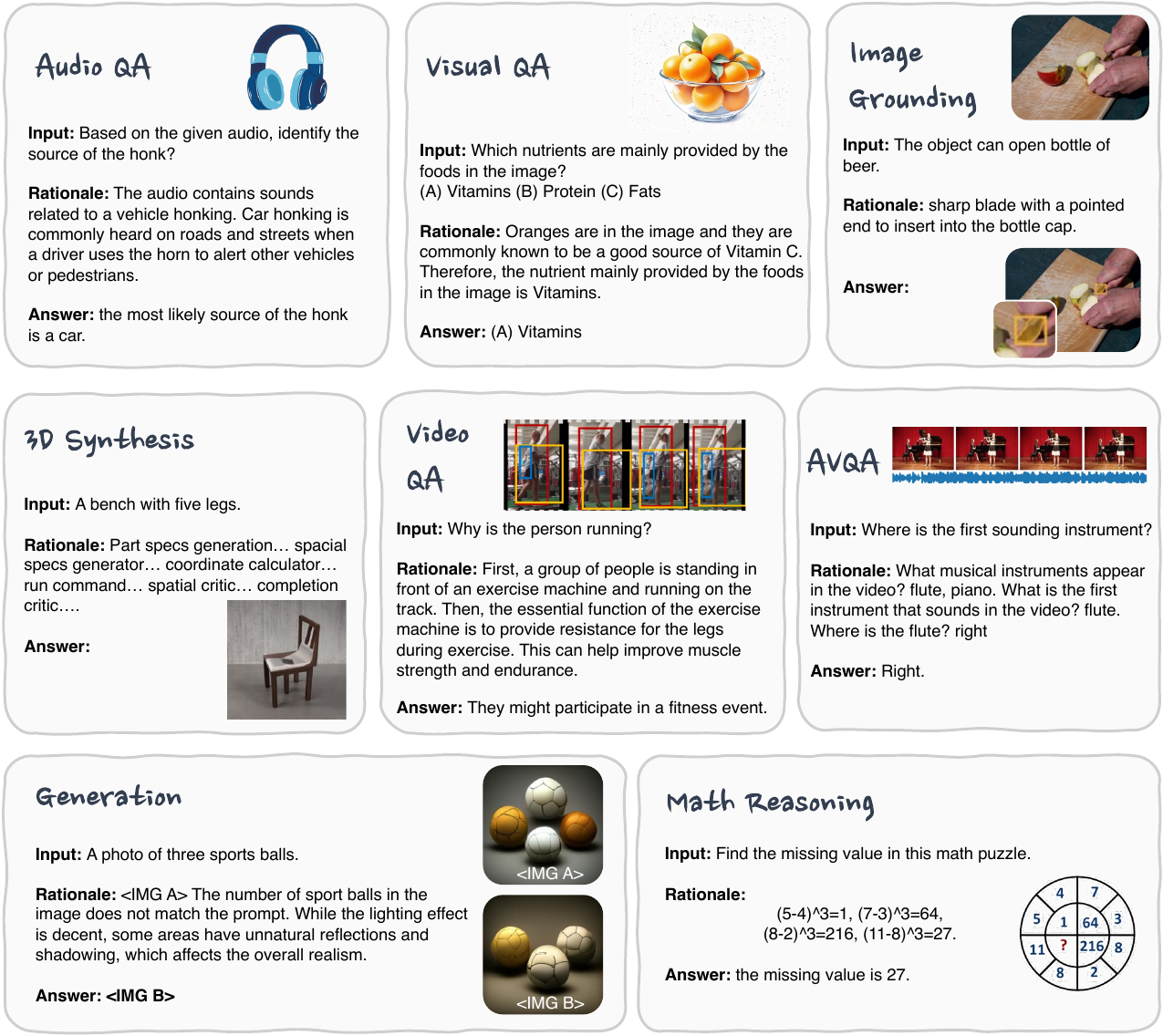}
    \caption{Examples of MCoT applications in various modalities and tasks.}
    \label{fig:tasks_with_prompts}
\end{figure}

MCoT extends the reasoning capabilities of LLMs/MLLMs to tackle complex tasks across diverse modalities, \eg, images, videos, audio, 3D, tables/charts and beyond, by employing chained thought reasoning. 
As demonstrated in \cref{fig:tasks_with_prompts}, integrating MCoT with these modalities facilitates the implementation of numerous fundamental and significant applications. This section systematically reviews MCoT research across these modalities, emphasizing key advancements and their contributions to the development of multimodal reasoning.

\subsection{MCoT Reasoning over Image}
The prevalence of image data and associated tasks has driven the extensive application of MCoT mostly in Visual Question Answering (VQA). 
Early implementations, such as IPVR \cite{chen2023see} and Multimodal-CoT \cite{zhang2023multimodal}, establish the foundational MCoT framework by generating intermediate rationales before final predictions. 
Subsequent advancements have further refined this paradigm: MC-CoT \cite{tan2024boosting} integrates self-consistency \cite{wang2022cot_sc} with MCoT, employing word-level majority voting during training to enhance the quality of generated rationales. 
SoT \cite{aytes2025sketchofthought} leverages a router model to dynamically select reasoning paradigms (\ie, conceptual chaining, chunked symbolism, and expert lexicons) inspired by human cognitive strategies to enhance reasoning efficiency.
CoCoT \cite{zhang2024cocot} improves multi-image comprehension in MLLMs through similarity and difference analysis across inputs, while RelationLMM \cite{xie2025relationlmm} explicitly addresses object relationship modeling via task decomposition. 
HoT \cite{yao2023HoT} extends the Graph-of-Thought framework by introducing hyperedges to connect multiple reasoning nodes, thereby enhancing multimodal reasoning capabilities. 

Structured reasoning mechanisms have been proposed to enhance controllability and interpretability.
DDCoT \cite{zheng2023ddcot} and Socratic Questioning \cite{hu2025socratic} employ staged reasoning processes to systematically refine multimodal outcomes. Interaction methodologies between text and vision modalities also critically influence rationale generation. Chain-of-Spot \cite{liu2024chain}, TextCoT \cite{luan2024textcot}, and DCoT \cite{jia2024dcot} prioritize region-of-interest analysis to improve contextual understanding. RAGAR \cite{khaliq2024ragar} and Cantor \cite{gao2024cantor} integrate automated processes with low-level image attributes to strengthen reasoning, whereas KAM-CoT \cite{mondal2024kam} and PKRD-CoT \cite{luo2024pkrd} incorporate external knowledge bases, which are further augmented by graph-based techniques described in \cite{yao2023beyond} and \cite{zheng2024picture}.
The reliance of MCoT on annotated reasoning data has spurred research into automated data augmentation. G-CoT \cite{ma2024dolphins}, STIC \cite{deng2024stic}, PS-CoT \cite{li2025ps}, SNSE-CoT \cite{zheng2024enhancing}, Chain-of-Exemplar \cite{luo2024chain}, and R-CoT \cite{deng2024r} address this limitation by innovating methods to automate and enhance training data generation. 
In addition, static image feature extraction often results in inconsistencies when handling complex reasoning demands. 
To mitigate this, DPMM-CoT \cite{he2024multi} and LLavA-AURORA \cite{bigverdi2024perception} regenerate image features from latent space.
Beyond text-based rationales, recent approaches leverage multimodal rationales for comprehensive reasoning, for instance, Visual-CoT \cite{rose2023visual}, Chain-of-Image \cite{meng2023chain}, VisualSketchpad \cite{hu2024visual}, MVoT \cite{li2025imagine}, and Visualization-of-Thought \cite{wu2024mind} effectively process multimodal scenes with multimodal thoughts.

MCoT's applicability also extends beyond VQA to specialized domains. 
For fine-grained instance-level tasks, CoTDet \cite{tang2023cotdet}, Det-Cot \cite{wu2024dettoolchain}, and CPSeg \cite{li2024cpseg} demonstrate notable advancements. In image generation, PromptCoT \cite{yao2024promptcot} focuses refining the input prompts, PARM++ \cite{guo2025can} optimizes reward mechanisms, and LayoutLLM-T2I \cite{qu2023layoutllm} and CreatiLayout \cite{zhang2024creatilayout} employ text-based layout construction prior to synthesis, significantly improving output quality.

\subsection{MCoT Reasoning over Video}
Video understanding also relies on essential reasoning capabilities, as beyond the image understanding that require process static visual content and spatial relationships, videos present challenges of temporal dynamics, particularly in the case of long videos.
As a basic usage, CaVIR \cite{li2023intentqa} enhances intent question answering, which demands contextual and commonsense comprehension, by implementing a zero-shot MCoT approach. 
Similarly, VideoAgent \cite{wang2024videoagent} and HM-Prompt \cite{sun2024hallucination} employ zero-shot MCoT to improve long-video reasoning and reduce hallucinations.
AntGPT \cite{zhao2023antgpt} extends few-shot MCoT to action categorization in egocentric videos.
In generative tasks, DreamFactory \cite{xie2024dreamfactory} applies few-shot MCoT to produce consistent key frames for long-video synthesis.

For complex video comprehension, Video-of-Thought \cite{fei2024video} proposes a comprehensive five-stage framework: task and target identification, object tracking, action analysis, ranked question answering, and answer verification. This structured approach ensures thorough interpretation of video content. Likewise, CaRDiff \cite{tang2024cardiff} decomposes intricate video tasks into sub-components—caption generation, saliency inference, and bounding box production—to guide diffusion processes for salient object mask creation. 
R3CoT \cite{hong2025following} introduces a three-stage model (\ie, refining, retrieving, reasoning) tailored for video rumor detection, while Grounding-Prompter \cite{chen2023grounding} integrates global and local perception for temporal sentence grounding, localizing video moments based on linguistic queries. 
Efficiency in long-video analysis is addressed by frameworks such as VIP \cite{himakunthala2023let}, which prioritizes key frames and extracts critical features (\eg, focus, action, emotion, objects, background) to evaluate reasoning through attribute prediction for intermediate and future frames. 
Chain-of-Shot \cite{hu2025cos} further optimizes frame sampling by employing binary video summaries during training and assessing frame-task relevance for efficient inference.

MCoT’s utility also spans specialized domains, such as medical video analysis \cite{plini2024ti,dai2024interpretable} and affective computing \cite{lee2024analyzing_emotion,lei2024emotion,zhao2025r1omni}. Collectively, these advancements underscore MCoT’s role in decomposing complex video tasks, enhancing reasoning accuracy, and improving computational efficiency across diverse applications, marking a significant step forward in long-video understanding.

\subsection{MCoT Reasoning over 3D}
Reasoning in 3D scenes entails significant challenges due to the integration of complex, high-dimensional data, including shape, spatial relationships, and physical properties. 
Traditional approaches reliant on costing manual annotations and inflexible rules, prompting the adoption of MCoT to decompose intricate tasks into manageable and structured processes.

Several frameworks exemplify MCoT’s efficacy in 3D generation. 
3D-PreMise \cite{yuan20243d_PreMise} employs MCoT to direct LLMs in generating 3D shapes and programming parameters, streamlining object synthesis. 
Similarly, L3GO \cite{yamada2024Co3D_Thoughts} introduces Chain-of-3D-Thought, enabling 3D image generation through iterative trial-and-error and tool invocation within simulated environments, enhancing adaptability and precision.
Gen2Sim \cite{katara2024gen2sim} leverages MCoT to advance robot skill learning by generating 3D assets as MCoT inputs, subsequently prompting LLMs to produce task descriptions and reward functions. This approach reduces human intervention while facilitating scalable and diverse task acquisition in simulations. 

When meeting language instructions, CoT3DRef \cite{abdelrahman2023cot3dref} breaks down complex 3D grounding into interpretable steps. Meanwhile, 3D-CoT \cite{chen2025integrating} improves 3D vision-language alignment by integrating MCoT with a dataset that includes structural reasoning annotations, covering aspects such as shape recognition, functional inference, and causal reasoning. Together, these advancements highlight the crucial role of MCoT for efficiently addressing complex and compositional 3D tasks.

\subsection{MCoT Reasoning over Audio and Speech}
MCoT has been effectively extended to step-wise and manageable speech and audio processing, mitigating the gap between the waveform signal and language semantics.
CoT-ST \cite{du2024cot_st} breaks the speech translation into discrete stages of speech recognition and subsequent translation. 
\citet{xie2025leveraging} integrate automatic speech recognition and emotion detection prior to empathetic dialogue generation. 
Audio-CoT \cite{ma2025audio_cot} incorporates vanilla CoT into audio understanding and reasoning tasks. 
Furthermore, Audio-Reasoner \cite{xie2025audio} achieves the first long-MCoT reasoning by integrating a four-step structured reasoning framework (\eg, Planning, Captioning, Reasoning, Summary).
For the generative task, SpatialSonic \cite{sun2024spatial_audio} employs vanilla MCoT to derive pertinent attributes and captions, supporting the creation of a spatial audio generation. 
SpeechGPT-Gen \cite{zhang2024speechgpt} further introduces the Chain-of-Information-Generation approach, which systematically models semantic and perceptual information in sequential steps to facilitate natural speech generation. 
These developments highlight the adaptability and efficacy of MCoT in enhancing speech and audio processing, fostering more natural and  contextually responsive outcomes.

\subsection{MCoT Reasoning over Table and Chart}
LLMs demonstrate proficiency in document comprehension but face challenges with structured data, such as tables and charts, due to their intricate layouts and implicit patterns. 
LayoutLLM \cite{luo2024layoutllm} advances document understanding by integrating layout-aware pretraining at the document, region, and segment levels, employing vanilla MCoT to enhance processing. 
Similarly, \citet{dai2025multimodal} incorporate scene graphs to interpret charts, utilizing vanilla MCoT to mitigate hallucination in LLMs' responses. 
While these efforts provide coarse-level analysis, recent methodologies address such limitations by decomposing tasks into sequential, executable operations. 
TableGPT \cite{zha2023tablegpt} introduces the Chain-of-Command approach, leveraging command sets (\eg, SelectCondition and GroupBy) to systematically process tabular questions. 
\citet{wang2024chain_of_table} propose Chain-of-Table, enabling LLMs to dynamically generate requisite operations and arguments, reconstructing tables to retain only pertinent information. 
In contrast, ReFocus \cite{fu2025refocus} emulates human attention by producing visual thoughts through editing operations, such as adding highlights or masking regions in tables, thereby improving comprehension. 
These advancements collectively illustrate the efficacy of MCoT in navigating the complexities of structured data.

\subsection{Cross-modal CoT Reasoning}
CoT-like reasoning also excels in integrating multiple modalities beyond text and a single additional modality, \ie, cross-modal CoT reasoning. 
AVQA-CoT \cite{li2024avqa_cot} decomposes complex queries into simpler sub-questions, addressing audio-visual question answering (AVQA) sequentially through LLMs and pre-trained models.
Similarly, SegPref \cite{wang2024avs_cot} employs VLLMs to detect potential sounding objects within visual scenes, subsequently combining text rationales with mask decoders for audio-visual segmentation (AVS), thereby reducing over-reliance on visual features. 
In parallel, Chain-of-Empathy \cite{zhang2025CoEmpathetic} leverages vanilla MCoT alongside psychotherapeutic principles to enhance LLMs’ reasoning about human emotions, promoting empathetic responses. 
Likewise, MM-PEAR-CoT \cite{li2024multimodal_emotion} applies vanilla MCoT to analyze linguistic emotion, integrating it with audio and video inputs to improve multimodal emotion recognition and mitigate hallucinations. 
R1-Omni \cite{zhao2025r1omni} presents the first application of Reinforcement Learning with Verifiable Reward (RLVR) to an Omni-multimodal LLM in the context of emotion recognition.
Although cross-modal CoT reasoning predominantly relies on text-based rationales, these advancements in MCoT demonstrate superior performance across diverse downstream tasks.

\section{Methodologies in MCoT Reasoning}
\label{Methodologies in MCoT Reasoning}
 
To thoroughly investigate the robust reasoning capabilities of MCoT in multimodal contexts, the research community has developed a wide array of methods and strategies centered on MCoT. 
To provide a systematic and comprehensive analysis, we categorize these approaches from multiple perspectives: rationale construction, structural reasoning, information enhancing, objective granularity, multimodal rationale, and test-time scaling.

\begin{figure}[!t]
    \centering
    \includegraphics[width=1\textwidth]{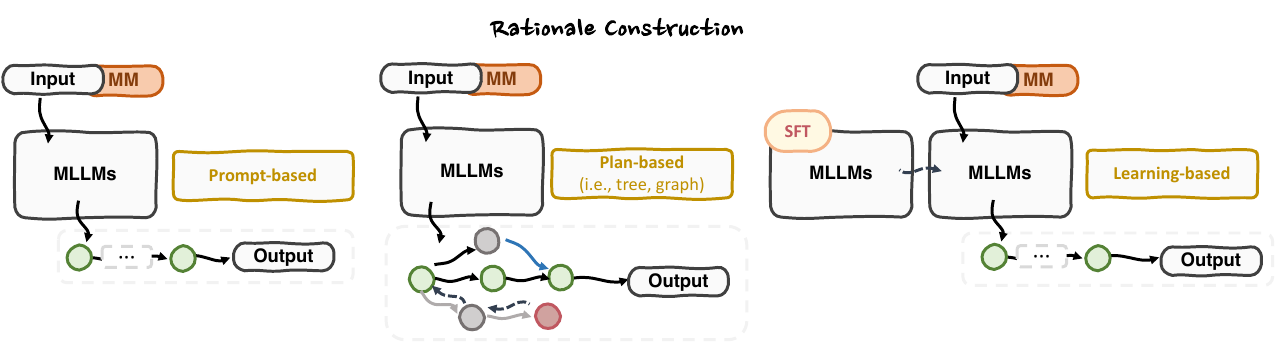}
    \caption{MCoT reasoning methods under different \textbf{\emph{rationale construction}} perspectives. 
    }
    \label{fig:method_x_reasoning_ability}
\end{figure}

\subsection{From Rationale Construction Perspective}
\label{sec:rationale_construction}
This part summarizes the methodologies employed in constructing rationales for MCoT reasoning. 
Unlike traditional direct input-output approaches, which prioritize final answers, CoT and MCoT emphasize deriving the correct answer through the reasoning process. 
Hence, MCoT reasoning methodologies primarily concern the construction of rationales and can be categorized into three distinct types: prompt-based, plan-based, and learning-based methods, as shown in \cref{fig:method_x_reasoning_ability}.

\paragraph{Prompt-based Methods.} 
Prompt-based MCoT reasoning employs carefully designed prompts, including instructions or in-context demonstrations, to guide models in generating rationales during inference, typically in zero-shot or few-shot settings. 
For instance, the simplest instruction is ``think step-by-step to understand the given text and image inputs'' serves as a zero-shot prompt \cite{himakunthala2023let} to elicit a rationale for addressing multimodal problems. 
However, most MCoT approaches specify explicit steps to ensure the reasoning follows specific guidance \cite{fei2024video,chen2023see,meng2023chain,luo2024pkrd,zheng2024thinking,wang2024videoagent,xie2025leveraging}. 
In addition, expert tools are often integrated to deepen insights into detailed information \cite{wu2024dettoolchain,gao2024cantor} or to incorporate multimodal data into verbal reasoning \cite{meng2023chain,hu2024visual,wu2024mind}, particularly in image and video understanding. 
In few-shot scenarios, prompts may include explicit reasoning examples to further steer the reasoning process \cite{zhao2023antgpt,chen2023grounding,plini2024ti}. 
This prompt-based methodology demonstrates notable flexibility, making it advantageous for scenarios where computational resources are constrained or swift responses are essential.

\paragraph{Plan-based Methods.}
Plan-based MCoT reasoning enables models to dynamically explore and refine thoughts during the reasoning process. 
MM-ToT \cite{gomez2023mmtot} utilizes GPT-4 \cite{achiam2023gpt4} and Stable Diffusion \cite{rombach2021stablediffusion} to generate multimodal outputs, applying DFS and BFS to select optimal outputs based on a 0.0–1.0 metric scale. 
HoT \cite{yao2023HoT} produces interconnected thoughts from multimodal inputs, encapsulated within a single hyperedge. 
In contrast, Aggregation Graph-of-Thought (AGoT) \cite{yang2024AGoT} constructs a reasoning aggregation graph, integrating multiple reasoning facets at each reasoning step and subsequently incorporating visual data. 
Blueprint Debate on Graph (BDoG) \cite{zheng2024picture} adopts a distinctive approach, forgoing search algorithms in favor of three agents—an affirmative debater, a negative debater, and a moderator. 
Through iterative debate, these agents address multimodal questions, with the moderator synthesizing a final answer, implicitly forming a graph-of-thought that explores and aggregates diverse thoughts. 
PARM++ \cite{guo2025can} trains an image generation model with chained verification steps, such as potential assessment, to filter out bad outputs during the image generation.
In summary, unlike prompt-based methods with their linear, example-driven inference, plan-based MCoT variants enable models to traverse multiple reasoning pathways, enhancing adaptability and problem-solving depth.

\paragraph{Learning-based Methods.}
Learning-based MCoT reasoning embeds rationale construction within the training or fine-tuning process, requiring models to explicitly learn reasoning skills alongside multimodal inputs. 
Multimodal-CoT \cite{zhang2023multimodal} pioneers this approach by fine-tuning models with reasoning data containing rationales, fostering inherent reasoning capabilities. 
PCoT \cite{wang2024t} refines this paradigm for rationale generation, while MC-CoT \cite{tan2024boosting} incorporates multimodal consistency and majority voting during training to enhance reasoning in smaller models. 
G-CoT \cite{ma2024dolphins} employs ChatGPT to produce reasoning data, activating reasoning potential transferable to autonomous driving via fine-tuning. 
LoT \cite{zhong2024let} boosts creativity through fine-tuning with leap-of-thought data, and PromptCoT \cite{yao2024promptcot} enhances prompt generation for image synthesis through targeted fine-tuning. 
In summary, Learning-based methods focus on embedding reasoning patterns during training. 
However, following the release of OpenAI o1 \cite{jaech2024o1} in late 2024, interest has surged in augmenting long-CoT reasoning with scaled test-time computation \cite{dong2024insight,guo2025deepseekr1,snell2024test_time_scaling,llmsdemystifying} to tackle complex reasoning tasks, which we will further discuss in the \cref{sec:o1_like}.

\begin{figure}[t!]
    \centering
    \includegraphics[width=1\textwidth]{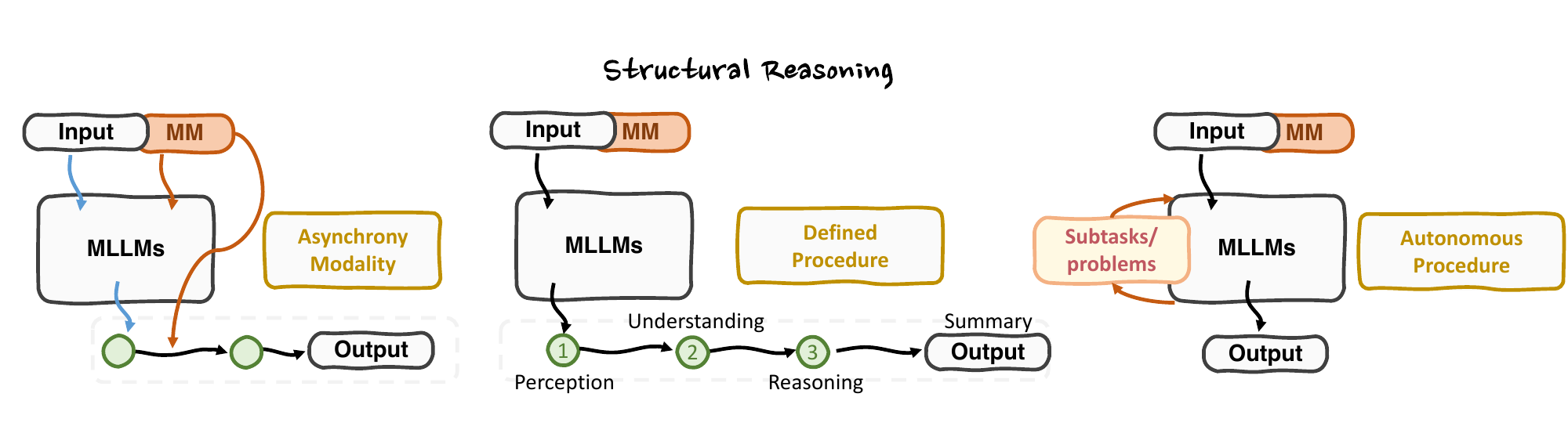}
    \caption{MCoT methods under different \textbf{\emph{structural reasoning}} perspectives. 
    }
    \label{fig:method_x_structural}
\end{figure}
 
\subsection{From Structural Reasoning Perspective}
Beyond simplistic rationale generation based on next token prediction, recent studies have proposed structural reasoning frameworks to enhance the controllability and interpretability of the rationale generation process. 
\cref{fig:method_x_structural} demonstrates the structured formats categorized into three types: asynchronous modality modeling, defined procedure staging, and autonomous procedure staging.

\paragraph{Asynchronous Modality Modeling.}
Early research explorations in MCoT generate rationales directly from multimodal contexts, as exemplified by Multimodal-CoT \cite{zhang2023multimodal}, Audio-CoT \cite{ma2025audio_cot} and Grounding-Prompter \cite{chen2023grounding}. 
However, neuroscience research by \citet{wu2023role} suggest that recognition and reasoning operate in distinct cognitive modules, adhering to a \textit{\emph{``description then decision''}} strategy. 
This insight has motivated asynchronous approaches to modality processing. For instance, IPVR \cite{chen2023see} introduced a three-stage ``see, think, confirm'' framework for VQA, decoupling perception from reasoning. 
Visualization-of-Thought \cite{wu2024mind} simulates mental imagery by generating 2D grid-based text representations to guide search and navigation tasks. 
Similarly, TextCoT \cite{luan2024textcot} employs a two-phase process: first summarizing the image context, then generating responses grounded in visual inputs. Cantor \cite{gao2024cantor} separates perception and decision-making stages, where the perception phase extracts low-level attributes (\eg, objects, colors, shapes) from images or textual descriptions, while the decision phase integrates these features for accurate problem-solving. 
In contrast, VIC \cite{zheng2024thinking} decomposes tasks into text-based sub-steps before incorporating visual inputs to derive final rationales. 
These methods align with neuroscience by isolating perceptual encoding from high-level reasoning, thereby enhancing interpretability and alignment with human cognitive processes.

\paragraph{Defined Procedure Staging.}
Several studies explicitly define structured reasoning stages to enhance process interpretability. 
BDoG \cite{zheng2024picture} employs a fixed debate-summarization pipeline with specialized agents, while Det-CoT \cite{wu2024dettoolchain} formalizes VQA reasoning into templated instruction parsing, subtask decomposition, execution, and verification. 
VisualSketchpad \cite{hu2024visual} structures rationales into ``Thought, Action, Observation'' phases, whereas CoTDet \cite{tang2023cotdet} implements object detection through object listing, affordance analysis, and visual feature summarization. 
Socratic Questioning \cite{hu2025socratic} decomposes VQA into self-guided subquestion generation, detailed captioning, and summarization.
The Grounding-Prompter \cite{chen2023grounding} performs global understanding, noise assessment, and partition understanding before the final decision-making. 
For multi-image comprehension, CoCoT \cite{zhang2024cocot} systematically compares similarities and differences across inputs.
LLaVA-CoT \cite{xu2024llava_cot} achieves long-MCoT reasoning via summary, captioning, analysis, and conclusion phases, and Audio-Reasoner \cite{xie2025audio} also implements in the same manner. 

Structured staging also facilitates dataset construction and downstream applications development. 
URSA \cite{luo2025ursa} generates mathematical reasoning datasets via rationale distillation and trajectory rewriting.
Paralleled by Chain-of-Sentiment \cite{luo2024panosent}, Chain-of-Exemplar \cite{luo2024chain} and Chain-of-Empathetic \cite{zhang2025CoEmpathetic} in educational and affective computing domains. 
SmartAgent \cite{zhang2024CoUser} builds personal assistants through GUI navigation, reasoning, and recommendation stages. 
CoT-ST \cite{du2024cot_st} combines speech recognition and machine translation for speech translation.
SegPref \cite{wang2024avs_cot} robustly locates the sounding objects in the visual space by leveraging global understanding, sounding object filtering, and noise information removal.
In generative tasks, PARM \cite{guo2025can} generates images with clarified judgment, potential assessment, and best-of-N selection, while SpeechGPT-Gen \cite{zhang2024speechgpt} synthesizes speech from a perceptual to a semantic perspective.

\paragraph{Autonomous Procedure Staging.}
Recent studies have explored autonomous procedural staging, enabling LLMs to self-determine the sequence of reasoning steps. PS-CoT \cite{li2025ps} allows LLMs to autonomously generate task-solving plans before rationale generation, while DDCoT \cite{zheng2023ddcot} and AVQA-CoT \cite{li2024avqa_cot} decompose problems into subquestions for iterative resolution. CoT-PT \cite{ge2023chain} employs hierarchical reasoning from abstract to concrete concepts (\eg, object $\rightarrow$ animal $\rightarrow$ dog). Image-of-Thought \cite{zhou2024image} automatically segments VQA tasks into subtasks with corresponding image manipulation actions. Insight-V \cite{dong2024insight} dynamically determines the focus of each reasoning step and autonomously decides whether to proceed or summarize intermediate results. Chain-of-Table \cite{wang2024chain_of_table} generates stepwise queries to modify table structures (\eg, adding a ``country'' header), synthesizes operation arguments, and optimizes data storage for efficient answer derivation. In embodied intelligence tasks, E-CoT \cite{zawalski2024robotic} and Emma-X \cite{sun2024emmax} enable LLMs to infer executable subtask sequences.

\subsection{From Information Enhancing Perspective}
Enhancing multimodal inputs facilitates comprehensive reasoning through the integration of expert tools and internal or external knowledge.

\paragraph{Using Expert Tools.} 
Recent studies leverage specialized tools to enhance multimodal reasoning through structured visual or geometric operations. For mathematical and geometric tasks, approaches such as Chain-of-Image \cite{meng2023chain} and VisualSketchpad \cite{hu2024visual} generate auxiliary visualizations via expert tools or codes. Similarly, Det-CoT \cite{wu2024dettoolchain}, Cantor \cite{gao2024cantor}, and Image-of-Thought \cite{zhou2024image} employ image manipulation tools (\eg, zoom-in, ruler markers) to improve fine-grained visual analysis. In parallel, L3GO \cite{yamada2024Co3D_Thoughts} and 3D-Premise \cite{yuan20243d_PreMise} integrate 3D generation tools to support spatial reasoning workflows. These methodologies underscore the growing emphasis on integrating domain-specific toolkits to augment both interpretability and precision in multimodal reasoning tasks.

\paragraph{Using World Knowledge Retrieval.}
Recent studies augment reasoning processes by integrating external knowledge sources. Approaches such as RAGAR \cite{khaliq2024ragar}, AR-MCTS \cite{dong2024progressive_retrieval}, and Chain-of-Action \cite{pan2024chain} leverage retrieval-augmented generation (RAG) to incorporate domain-specific or commonsense knowledge during inference. G-CoT \cite{ma2024dolphins} distills task-relevant commonsense information from ChatGPT, while CoTDet \cite{tang2023cotdet} retrieves object affordances to provide context for detection tasks. KAM-CoT \cite{mondal2024kam} jointly reasons over images, textual data, and structured knowledge graphs to enhance multimodal comprehension. These methods demonstrate the critical role of knowledge-aware architectures in bridging perceptual inputs with conceptual understanding.

\paragraph{Leveraging In-context Knowledge Retrieval.}
Beyond external knowledge augmentation, several studies improve reasoning by retrieving and organizing information directly from input content or generated rationales from LLMs/MLLMs themselves. DCoT \cite{jia2024dcot} focuses on prioritizing image regions of interest during inference. In contrast, MCoT-Memory \cite{liang2024memory_driven}, MGCoT \cite{yao2023beyond}, Video-of-Thought \cite{fei2024video}, CCoT \cite{mitra2024compositional}, and BDoG \cite{zheng2024picture} implicitly retrieve in-context knowledge by modeling relationships between objects or concepts through scene graph representations. Similarly, CoT3DRef \cite{abdelrahman2023cot3dref} generates target anchors in grounding reference sentences, effectively functioning as simplified scene graphs. Together, these approaches demonstrate the effectiveness of structured in-context knowledge extraction in enhancing reasoning fidelity.

\begin{figure}[t!]
    \centering
    \includegraphics[width=1\textwidth]{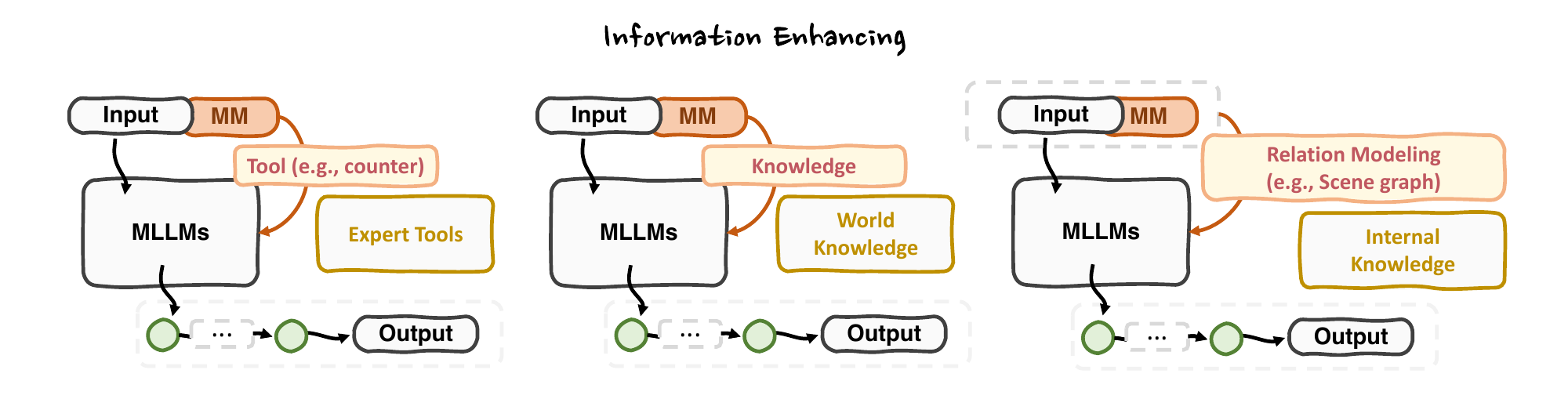}
    \caption{MCoT reasoning under perspectives with \textbf{\emph{information enhancing}}.
    }
    \label{fig:method_x_info_enhance}
\end{figure}

\subsection{From Objective Granularity Perspective}
\begin{figure}[t!]
    \centering
    \includegraphics[width=1\textwidth]{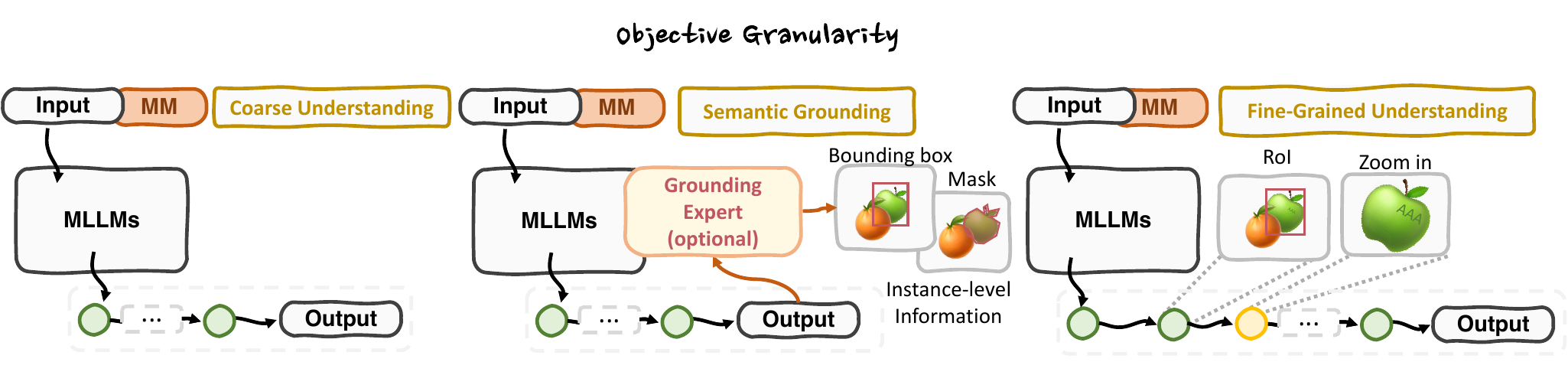}
    \caption{MCoT reasoning under the perspectives of various \textbf{\emph{objective granularities}}.
    }
    \label{fig:method_x_task_granularity}
\end{figure}

Research methodologies often align with the granularity of objective, as illustrated in \cref{fig:method_x_task_granularity}. While most question-answering tasks emphasize overview information, \ie, coarse understanding, some fine-grained tasks such as grounding place greater importance on individual instances, \ie, semantic grounding and fine-grained understanding.

\paragraph{Coarse Understanding Level.} 
As the most widely explored information processing level, coarse understanding is commonly used in tasks such as VQA and AQA, exemplified by methods like Multimodal-CoT \cite{zhang2023multimodal} and Audio-CoT \cite{ma2025audio_cot}. These approaches aim to achieve an overview of the given multimodal information without focusing on the details.

\paragraph{Semantic Grounding Level.} 
Semantic grounding tasks are addressed through specialized reasoning paradigms. CPSeg \cite{li2024cpseg}, CoTDet \cite{tang2023cotdet}, and Migician \cite{li2025migician} refine grounding references via LLMs to enhance the alignment between textual prompts and target visual instances, improving precision in downstream mask decoders or bounding box proposers. Similarly, SegPref \cite{wang2024avs_cot} utilizes Visual Large Language Models (VLLMs) to infer potential sounding objects from global scene information, then incorporates audio information to locate the sounding objects within the visual space.

\paragraph{Fine-grained Understanding Level.}
Fine-grained understanding also necessitates capturing detailed information within multimodal contexts. DCoT \cite{jia2024dcot}, TextCoT \cite{luan2024textcot}, and Chain-of-Spot \cite{liu2024chain} first focus on regions of interest within images, followed by the identification of fine-grained information from these retrieved areas. E-CoT \cite{zawalski2024robotic} retrieves bounding boxes of target objects and identifies gripper positions for robotic tasks, facilitating embodied interactions.

\begin{figure}[t!]
    \centering
    \includegraphics[width=1\textwidth]{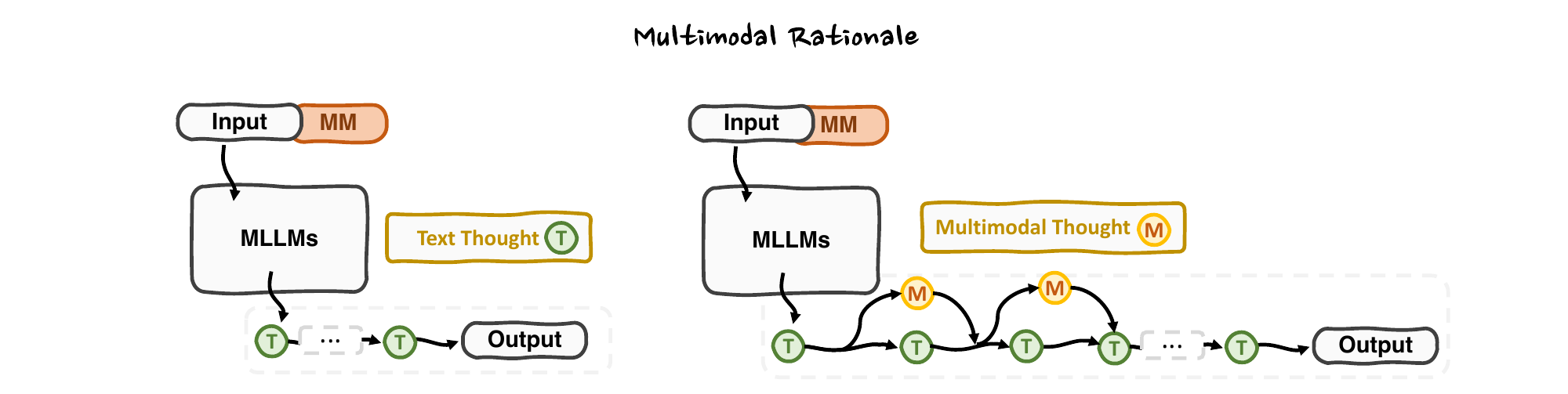}
    \caption{MCoT reasoning with \textbf{\emph{multimodal rationale}}.
    }
    \label{fig:method_x_mm_rationale}
\end{figure}

\subsection{From Multimodal Rationale Perspective}
As introduced in the preliminary Section \S\ref{Background and Preliminary}, reasoning processes can adopt either text-only or multimodal rationales. 
The predominant focus has been on text-centric approaches, exemplified by Multimodal-CoT \cite{zhang2023multimodal}, PCoT \cite{wang2024t}, MC-CoT \cite{wei2024mccot}, LLaVA-CoT \cite{xu2024llava_cot}, and Grounding-Prompter \cite{chen2023grounding}. 
These approaches predominantly employ textual representation to encode multimodal information, enabling seamless integration with the reasoning mechanisms of LLMs or MLLMs.

Emerging methods, however, explore multimodal rationale construction inspired by human cognitive processes. For instance, ReFocus \cite{fu2025refocus} simulates visual attention by overlaying highlighted regions on tabular data, while Visual-CoT \cite{rose2023visual} addresses logical gaps in sequential image reasoning by generating intermediate imaginary states. Chain-of-Image \cite{meng2023chain} emulates human tool-assisted reasoning by generating auxiliary diagrams for mathematical or geometric problem-solving, and Image-of-Thought \cite{zhou2024image} combines textual and visual retrieval to dynamically localize objects which should be referenced to answer a fine-grained question. 
Visualization-of-Thought \cite{wu2024mind} constructs 2D grids to represent human mental images for solving spatial reasoning tasks, whereas MVoT \cite{li2025imagine} further visualizes each reasoning step. 
Similarly, CoTDiffusion \cite{ni2024generate} leverages diffusion models to decompose robotic manipulation tasks into coherent visual subgoal plans, bridging abstract reasoning and physical execution.
This progression from text-centric to multimodal rationales reflects an increasing emphasis on emulating human-like cognitive mechanisms.

\subsection{From Test-Time Scaling Perspective}
 
\label{sec:o1_like}

The reasoning response of LLMs can be categorized into direct responses and CoT responses, analogous to the two distinct reasoning systems present in human cognition \cite{evans2003system2}.
System 1, characterized by rapid, heuristic-driven decision-making, contrasts with System 2, which employs deliberate, logical reasoning to yield more accurate and less biased outcomes \cite{li2025system}. 
\citet{snell2024test_time_scaling} further substantiated the efficacy of ``slow thinking'' in LLMs that, based on System-2, scaling test-time computation optimally during inference may outperform scaling model parameters in efficiency.
The release of OpenAI's o1 \cite{jaech2024o1} has further fueled interest in large-scale reasoning models that combine both internal and external slow-thinking mechanisms \cite{jiang2024technical,min2024imitate,gan2025rethinking,chen2025towards}, offering potential solutions to complex challenges, particularly in fields like mathematics and coding, while Deepseek-R1 \cite{guo2025deepseekr1} showcases that reinforcement learning (RL) alone can awaken the long-CoT reasoning ability.

\paragraph{Slow-Thinking-based Models.}
\textit{\emph{Internal slow-thinking}} enhances reasoning depth and quality through training or finetuning. 
Conversely, \textit{\emph{external slow-thinking}} improves reasoning with iterative sampling and refining solutions during inference. 
As a pioneering work, Qwen-QwQ \cite{qwen2024qwq} undergoes supervised fine-tuning (SFT) with 7,000 long CoT samples, unlocking the long-CoT reasoning capability.
Macro-o1 \cite{zhao2024marco} also integrates internal slow-thinking via CoT fine-tuning but further employs Monte Carlo Tree Search (MCTS), a heuristic search algorithm, to activate the external slow-thinking capability. 

Building on these foundations, research has extended into large multimodal reasoning models. 
Visual-o1 \cite{ni2024visual_o1} addresses ambiguous instructions using a multimodal and multi-turn CoT framework, while LLaVA-CoT \cite{xu2024llava_cot} and Audio-Reasoner \cite{xie2025audio} achieve long-McoT reasoning with SFT and structural reasoning. 
LlamaV-o1 \cite{thawakar2025llamav_o1} active the long-MCoT reasoning ability through a curriculum learning approach. 
Virgo \cite{du2025virgo} constructs a multimodal slow-thinking system by fine-tuning an MLLM with a compact set of long-form textual data.
AR-MCTS \cite{dong2024progressive_retrieval} dynamically retrieves multimodal insights during MCTS expansion to enrich sampling diversity and reliability.
AStar \cite{wu2025boosting} distills reasoning patterns from only 500 data samples via MCTS to steer reasoning, whereas Mulberry \cite{yao2024mulberry} enhances tree search with collective learning from multiple MLLMs, leveraging negative paths to generate reflective data for improved self-reflection. 
RedStar \cite{xu2025redstar} demonstrates that a few thousand samples suffice to activate long-CoT capabilities, with efficacy scaling alongside model size, and can enhance the generalization capability of MLLMs.
These developments underscore the transformative potential of slow-thinking paradigms in advancing multimodal reasoning capabilities.

\begin{figure}[t!]
    \centering
    \includegraphics[width=1\textwidth]{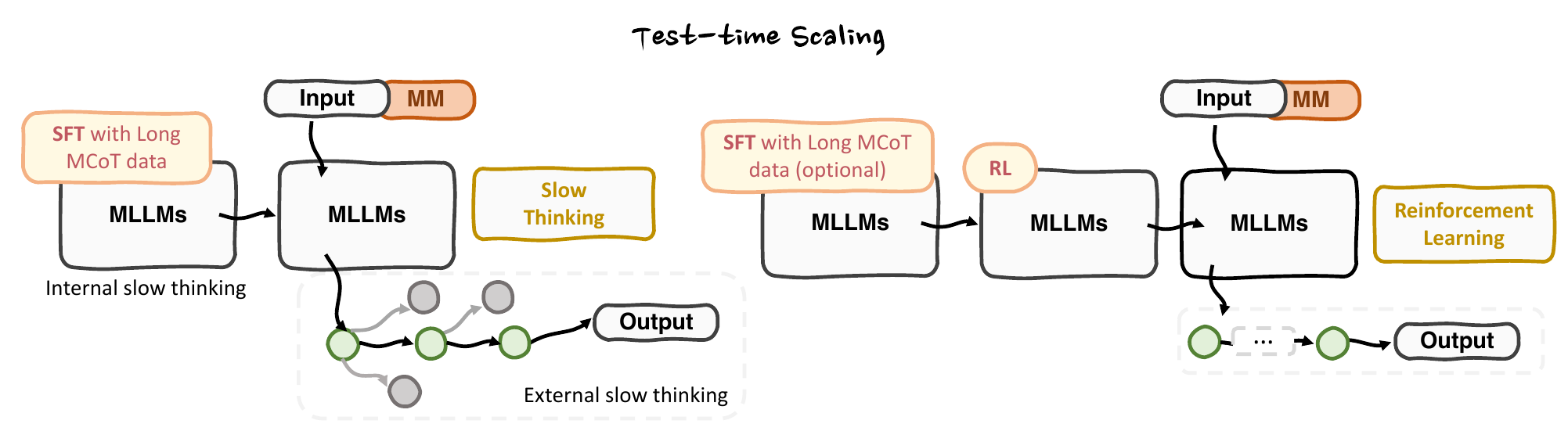}
    \caption{MCoT reasoning with \textbf{\emph{test-time scaling}} strategies. 
    RL can help improve reasoning quality, or active long-CoT reasoning ability without annotated long-CoT training data. SFT is optional.
    }
    \label{fig:method_x_scaling}
\end{figure}

\paragraph{Reinforcement Learning-based Models.}
RL has demonstrated significant efficacy in advancing the reasoning capabilities of LLMs. 
Deepseek-R1 \cite{guo2025deepseekr1} illustrates this by activating long-CoT reasoning using RL alone, with further improvements surpassing OpenAI o1 in some aspects via SFT cold starts and iterative self-improvement, sparking interest in RL-driven models like Open-R1 \cite{openr1} and TinyZero \cite{tinyzero}.
In multimodal contexts, early efforts like LLaVA-Reasoner \cite{zhang2024llavareasoner} and Insight-V \cite{dong2024insight} refined reasoning capabilities by employing long-MCoT data fine-tuning and direct preference optimization (DPO) guided by human feedback. 
Then advancing based on the Deepseek-R1, Multimodal-Open-R1 \cite{multimodal-open-r1} integrates the GPRO framework \cite{guo2025deepseekr1} to develop an R1-like MLLM, while R1-V \cite{chen2025r1v} verifies that RL can enhance the generalization in visual reasoning tasks. 
The effectiveness of RL in visual reasoning is also verified by concurrent works such as R1-OneVision \cite{R1-OneVision}, VLM-R1 \cite{shen2025vlmr1}, LMM-R1 \cite{peng2025lmmr1}, and Easy-R1 \cite{zheng2025easyr1}.
In addition, compared to the aforementioned outcome reward models (ORM), progress reward models (PRM) such as MSTaR \cite{liu2024diving} and VisualPRM \cite{wang2025visualprm} assess and provide feedback at each reasoning step, further enhancing the self-consistency and self-evolving capabilities of MLLMs.

This RL-based reasoning paradigm further extends to video reasoning tasks with Open-R1-Video \cite{wang-2025-open-r1-video}, detection tasks with Curr-ReFT \cite{deng2025vlmreasoning}, segmentation tasks with Seg-Zero \cite{liu2025segzero}, and multimodal emotion recognition tasks incorporating audio in R1-Omni \cite{zhao2025r1omni}.
Moreover, RL fosters the emergence of ``aha-moments'' that enables the reflection and backtracking during the reasoning, which is first identified by Deepseek-R1 in text-only scenarios. MM-Eureka \cite{meng2025mmeureka} and VisualThinker-R1-Zero \cite{zhou2025VisualThinker-R1-Zero} further reproduce the phenomena successfully in visual reasoning. 
\cref{tab:rl_models} concludes the techniques used by MLLMs with RL for better long-MCoT reasoning. 
In summary, RL unlocks complex reasoning and ``aha-moment'' without SFT, demonstrating its potential to enhance model capabilities through iterative self-improvement and rule-based approaches, ultimately paving the way for more advanced and autonomous multimodal reasoning systems.

\begin{table}[!t]
\fontsize{8.5}{11}\selectfont
\centering
\setlength{\tabcolsep}{0.6mm}
\begin{tabular}{lcccccc}
\toprule
\rowcolor{headcolor}
\textbf{Model} & \textbf{Foundational LLMs} & \textbf{Modality} & \textbf{Learning} & \textbf{Cold Start} & \textbf{Algorithm} & \textbf{Aha-moment} \\
\midrule
Deepseek-R1-Zero \cite{guo2025deepseekr1} & Deepseek-V3 & T & RL & \ding{55} & GRPO & \ding{52} \\
\rowcolor{tablegrey} Deepseek-R1 \cite{guo2025deepseekr1}  & Deepseek-V3 & T & SFT+RL & \ding{52} & GRPO & - \\ \hline
LLaVA-Reasoner \cite{zhang2024llavareasoner} & LLaMA3-LLaVA-NEXT-8B & T,I & SFT+RL & \ding{52} & DPO & - \\
\rowcolor{tablegrey} Insight-V \cite{dong2024insight} & LLaMA3-LLaVA-NEXT-8B & T,I & SFT+RL & \ding{52} & DPO & - \\ \hline
Multimodal-Open-R1 \cite{multimodal-open-r1} & Qwen2-VL-7B-Instruct & T,I & RL & \ding{55} & GRPO & \ding{55} \\
\rowcolor{tablegrey} R1-OneVision \cite{R1-OneVision} & Qwen2.5-VL-7B-Instruct & T,I & SFT & - & - & - \\
R1-V \cite{chen2025r1v} & Qwen2.5-VL & T,I & RL & \ding{55} & GPRO & \ding{55} \\
\rowcolor{tablegrey} VLM-R1 \cite{shen2025vlmr1} & Qwen2.5-VL & T,I & RL & \ding{55} & GPRO & \ding{55} \\
LMM-R1 \cite{peng2025lmmr1} & Qwen2.5-VL-Instruct-3B & T,I & RL & \ding{55} & PPO & \ding{55} \\
\rowcolor{tablegrey} Curr-ReFT \cite{deng2025vlmreasoning} & Qwen2.5-VL-3B & T,I & RL+SFT & \ding{55} & GPRO & - \\
Seg-Zero \cite{liu2025segzero} &  Qwen2.5-VL-3B + SAM2 & T,I & RL & \ding{55} & GPRO & \ding{55} \\
\rowcolor{tablegrey} MM-Eureka \cite{meng2025mmeureka} & InternVL2.5-Instruct-8B & T,I & SFT+RL & \ding{52} & RLOO & - \\
MM-Eureka-Zero \cite{meng2025mmeureka} & InternVL2.5-Pretrained-38B & T,I & RL & \ding{55} & RLOO & \ding{52} \\
\rowcolor{tablegrey} VisualThinker-R1-Zero \cite{zhou2025VisualThinker-R1-Zero} & Qwen2-VL-2B & T,I & RL & \ding{55} & GPRO & \ding{52} \\
Easy-R1 \cite{zheng2025easyr1} & Qwen2.5-VL & T,I & RL & \ding{55} & GRPO & - \\
\rowcolor{tablegrey} Open-R1-Video \cite{wang-2025-open-r1-video} & Qwen2-VL-7B & T,I,V & RL & \ding{55} & GRPO & \ding{55} \\
R1-Omni \cite{zhao2025r1omni} & HumanOmni-0.5B & T,I,V,A & SFT+RL & \ding{52} & GRPO & - \\ 
\rowcolor{tablegrey} VisRL \cite{chen2025visrl} & Qwen2.5-VL-7B & T,I & SFT+RL & \ding{52}  & DPO & -\\
R1-VL \cite{zhang2025r1vl} & Qwen2-VL-7B & T,I & RL & \ding{55} & StepGRPO & - \\
\bottomrule
\end{tabular}%
\vspace{1mm}
\caption{Multimodal reasoning models utilizing reinforcement learning. Deepseek-R1 serves as a text-only LLM for comparison.}
\label{tab:rl_models}
\end{table}
 
\section{Applications with MCoT Reasoning} 
\label{sec:applications}
MCoT's robust capability to decompose complex tasks into manageable subtasks has facilitated its application across diverse domains, including embodied systems, agents, autonomous driving, healthcare innovations, and multimodal generation frameworks. 
Each domain exemplifies how MCoT reasoning enhances task decomposition, decision-making, and generalization, thereby offering significant insights into its transformative potential in addressing real-world AI challenges.

\subsection{Embodied AI}
Recent advancements in embodied AI have significantly enhanced robotic capabilities across planning, manipulation, and navigation. EmbodiedGPT \cite{mu2023embodiedgpt} and E-CoT \cite{zawalski2024robotic} utilize MCoT reasoning to segment tasks into actionable subgoals. Notably, EmbodiedGPT introduces the EgoCoT dataset for vision-language pre-training, while E-CoT focuses on the sequential execution of textual commands.
ManipLLM \cite{li2024manipllm} enhances manipulation via fine-tuned MLLMs for object-centric tasks, while CoTDiffusion \cite{ni2024generate} employs diffusion-generated visual subgoals to achieve precision in long-horizon activities. 
In spatial reasoning, Emma-X \cite{sun2024emmax} integrates grounded planning and predictive movement, while SpatialCoT \cite{liu2025spatialcot} uses coordinate alignment for complex spatial reasoning. 
In navigation, MCoCoNav \cite{shen2024enhancing} optimizes multi-robot coordination with a global semantic map and score-based collaboration, whereas MCoT-Memory \cite{liang2024memory_driven} improves long-horizon planning by incorporating memory retrieval and scene graph updates, retaining high-confidence experiences for robust decision-making. 
Collectively, these studies underscore a trend toward integrating multimodal data and chained reasoning for adaptable, generalizable embodied systems.

\subsection{Agentic System}
Advancements in AI-driven agent systems have expanded autonomous interaction and content generation capabilities. 
Auto-GUI \cite{zhang2023CoAction} employs Multimodal Chain-of-Action (MCoA) to manipulate graphical interfaces directly, enhancing efficiency without reliance on external tools or APIs. 
Similarly, SmartAgent \cite{zhang2024CoUser} integrates GUI navigation with Chain-of-User-Thought (CoUT) reasoning to provide personalized recommendations for embodied agents. 
In video understanding, VideoAgent \cite{wang2024videoagent} leverages LLMs with a reflective three-step decision process for accurate interpretation of long-form content. 
Complementing these, DreamFactory \cite{xie2024dreamfactory} pioneers long-video generation through a multi-agent framework, ensuring scene consistency via keyframe iteration and MCoT reasoning. 
These studies collectively illustrate the pivotal role of chaining mechanisms and agent collaboration in addressing complex real-world AI challenges.

In addition, a significant paradigm shift in AI agent systems has emerged recently, integrating ``perception-reasoning'' with ``planning-execution.'' 
The advent of Manus \cite{manus2025} exemplifies this transition, igniting interest in tool-use AI agents like OpenManus \cite{openmanus2025}. 
Leveraging LLMs for natural language understanding and generation, Manus iteratively refines solutions through goal-directed self-reflection. 
As a tool-use agent, it incorporates diverse functionalities, such as web search, data querying, and code execution, employing a chain-of-tools approach to address complex multimodal tasks in the real world. 
Future advancements in tool-use agents are expected to build upon foundational models, enhancing long-MCoT reasoning and integrating varied multimodal interfaces and tools. 
This developmental trajectory suggests a progression toward AI agents with increasingly human-like capabilities.

\subsection{Autonomous Driving}
Recent advancements in autonomous driving have increasingly leveraged MLLMs and MCoT reasoning to improve decision-making and adaptability. 
DriveCoT \cite{wang2024drivecot} integrates MCoT into end-to-end driving systems, supported by a tailored dataset, while PKRD-CoT \cite{luo2024pkrd} employs zero-shot MCoT prompting to address perception, knowledge, reasoning, and decision-making in dynamic environments. 
Human interaction is emphasized by \citet{ma2024learning} and \citet{cui2024receive}, who incorporate LLMs to interpret feedback and verbal instructions effectively. 
Sce2DriveX \cite{zhao2025sce2drivex} enhances end-to-end control through multimodal scene understanding and demonstrates robust generalization.
Furthermore, Reason2Drive \cite{nie2024reason2drive} provides 600K+ video-text pairs to explore interpretable reasoning, augmenting LLMs with object-level perception to strengthen planning capabilities.
Collectively, these efforts indicate a transition toward human-like reasoning, enhanced interactivity, and improved generalization within autonomous driving systems.

\subsection{Medical and Healthcare}
The innovative applications of AI in healthcare have harnessed chained reasoning to enhance various medical tasks. 
StressSelfRefine \cite{dai2024interpretable} detects stress in videos through a psychology-inspired ``Describe, Assess, Highlight'' process, refined via DPO to enhance accuracy. 
TI-PREGO \cite{plini2024ti} integrates ICL with Automatic Chain-of-Thought (ACoT) to identify procedural errors in egocentric videos, leveraging action sequences and logical reasoning. 
Chain-of-Look \cite{xi2023chainoflook} tackles surgical triplet recognition in endoscopic videos by breaking down tasks into video reasoning stages using vision-language prompts. 
Meanwhile, MedCoT \cite{liu2024medcot} improves medical visual question answering through a hierarchical expert system that culminates in a Mixture-of-Experts diagnosis. 
Furthermore, MedVLM-R1 \cite{pan2025medvlm} employs RL with only 600 medical VQA samples, aiming to enhance the medical reasoning capabilities of vision-language models.
Collectively, these efforts illustrate the effectiveness of MCoT reasoning in enhancing interpretability and precision across a range of medical AI applications.

\subsection{Social and Human}
MCoT has been effectively extended to the humanities and social sciences, leveraging their capacity for complex task decomposition. For instance, Chain-of-Empathetic \cite{zhang2025CoEmpathetic} employs MCoT to facilitate empathetic dialogue generation. 
MM-PEAR-CoT framework \cite{li2024multimodal_emotion} enhances multimodal sentiment analysis through a structured Preliminaries-Question-Answer-Reason approach, generating textual rationales prior to late-stage multimodal fusion. 
Advancing the affective computing domain, Chain-of-Sentiment \cite{luo2024panosent} refines sentiment analysis in conversational contexts, with complementary contributions from concurrent studies \cite{lee2024analyzing_emotion,lei2024emotion}. 
Furthermore, Chain-of-Exemplar \cite{luo2024chain} expands MCoT into the field of education,  X-Reflect \cite{lyu2024x} extends MCoT into multimodal recommendation systems, while \citet{yu2024chain} employs zero-shot MCoT for demographic inference. 
These advancements highlight the potential of MCoT to tackle complex challenges in human-centric and social scientific contexts.

\subsection{Multimodal Generation}
Recent advancements in AI-driven image and 3D generation highlight diverse multimodal synthesis strategies. 
PARM and PARM++ \cite{guo2025can} employ iterative step-by-step reasoning with clarity and potential assessments, augmented by reflection mechanisms, to produce high-quality images. 
GoT \cite{fang2025got} constructs a generation chain of thought by introducing a clear language reasoning process that analyzes semantic relationships and spatial arrangements before generating and editing images.
RPG-DiffusionMaster \cite{yang2024mastering} utilizes MLLMs for text-to-image diffusion, breaking down prompts into detailed subregions for coherent outputs, while L3GO \cite{yamada2024Co3D_Thoughts} leverages language agents with a Chain-of-3D-Thoughts approach to create unconventional 3D objects, outperforming traditional diffusion models on out-of-distribution descriptions. 
Additionally, 3D-PreMise \cite{yuan20243d_PreMise} integrates LLMs with program synthesis to generate parametric 3D shapes, yielding promising industrial outcomes when guided by explicit reasoning examples. 
These studies collectively underscore the potential of reasoning-augmented AI to overcome the limitations of data-driven generation, enabling precise and innovative multimodal outputs.

\begin{table}[!t]
\fontsize{8.5}{11}\selectfont
\centering
\setlength{\tabcolsep}{0.6mm}
\begin{tabular}{lcccccc}
\toprule
\rowcolor{headcolor}
\textbf{Datasets} & \textbf{Year} & \textbf{Task} & \textbf{Domain} & \textbf{Modality} & \textbf{Format} & \textbf{Samples} \\ \midrule
\multicolumn{7}{c}{\textbf{\textit{Training with rationale}}} \\
\rowcolor{tablegrey} ScienceQA \cite{lu2022scienceqa} & 2022 & VQA & Science & T, I & MC & 21K \\
A-OKVQA \cite{schwenk2022okvqa} & 2022 & VQA & Common & T, I & MC & 25K \\
\rowcolor{tablegrey} EgoCoT \cite{mu2023embodiedgpt} & 2023 & VideoQA & Common & T, V & Open & 200M \\
VideoCoT \cite{wang2024videocot} & 2024 & VideoQA & Human Action & T, V & Open & 22K \\
\rowcolor{tablegrey} VideoEspresso \cite{han2024videoespresso} & 2024 & VideoQA & Common & T, V & Open & 202,164 \\
EMMA-X \cite{sun2024emmax} & 2024 & Robot Manipulation & Indoor & T, V & Robot Actions & 60K \\
\rowcolor{tablegrey} M3CoT \cite{Chen0ZC0C24} & 2024 & VQA & Science, Math, Common & T, I & MC & 11.4K \\
MAVIS \cite{zhang2024mavis} & 2024 & ScienceQA & Math & T, I & MC and Open & 834K \\ 
\rowcolor{tablegrey} LLaVA-CoT-100k \cite{xu2024llava_cot} & 2024 & VQA & Common, Science & T, I & MC and Open & 834K \\ 
MAmmoTH-VL \cite{guo2024mammoth} & 2024 & Diverse & Diverse & T, I & MC and Open & 12M \\ 
\rowcolor{tablegrey} Mulberry-260k \cite{yao2024mulberry} & 2024 & Diverse & Diverse & T, I & MC and Open & 260K \\
MM-Verify \cite{sun2025mmverify} & 2025 & MathQA & Math & T, I & MC and Open & 59,772 \\
\rowcolor{tablegrey} VisualPRM400K \cite{wang2025visualprm} & 2025 & ScienceQA & Math, Science & T, I & MC and Open & 400K \\ 
R1-OneVision \cite{R1-OneVision} & 2025 & Diverse & Diverse & T, I & MC and Open & 155K \\ 

\hline
\multicolumn{7}{c}{\textbf{\textit{Evaluation without rationale}}} \\
\rowcolor{tablegrey} MMMU \cite{yue2024mmmu} & 2023 & VQA & Arts, Science & T, I & MC and Open & 11.5K \\
SEED \cite{li2024seed} & 2023 & VQA & Common & T, I & MC & 19K \\
\rowcolor{tablegrey} MathVista \cite{lu2023mathvista} & 2023 & ScienceQA & Math & T, I & MC and Open & 6,141 \\
MathVerse \cite{zhang2024mathverse} & 2024 & ScienceQA & Math & T, I & MC and Open & 15K \\
\rowcolor{tablegrey} Math-Vision \cite{wang2025mathvision} & 2024 & ScienceQA & Math & T, I & MC and Open & 3040 \\
OSWorld \cite{xie2024osworld} & 2024 & Agent & Real Comp. Env. & T,I & Agent Actions & 369  \\
\rowcolor{tablegrey} AgentClinic \cite{schmidgall2024agentclinic} & 2024 & MedicalQA & Medical & T,I & Open & 335  \\
MeViS \cite{ding2023mevis} & 2023 & Referring VOS & Common & T, V & Dense Mask & 2K \\
\rowcolor{tablegrey} VSIBench \cite{yang2024vsibench} & 2024 & VideoQA & Indoor & T, V & MC and Open & 5K \\
HallusionBench \cite{guan2024hallusionbench} & 2024 & VQA & Common & T, I & Yes-No & 1,129 \\
\rowcolor{tablegrey} AV-Odyssey \cite{gong2024avody} & 2024 & AVQA & Common & T, V, A & MC & 4,555 \\
AVHBench \cite{sung2024avhbench} & 2024 & AVQA & Common & T, V, A & Open & 5,816 \\
\rowcolor{tablegrey} RefAVS-Bench \cite{wang2024refavs} & 2024 & Referring AVS & Common & T, V, A & Dense Mask & 4,770 \\
MMAU \cite{sakshi2024mmau} & 2024 & AQA & Common & T, A & MC & 10K \\
\rowcolor{tablegrey} AVTrustBench \cite{chowdhury2025avtrustbench} & 2025 & AVQA & Common & T, V, A & MC and Open & 600K \\
MIG-Bench \cite{li2025migician} & 2025 & Multi-image Grounding & Common & T, I & BBox & 5.89K \\ 
\rowcolor{tablegrey} MedAgentsBench \cite{tang2025medagentsbench} & 2025 & MedicalQA & Medical & T, I & MC and Open & 862 \\

\hline
\multicolumn{7}{c}{\textbf{\textit{Evaluation with rationale}}} \\
\rowcolor{tablegrey} CoMT \cite{cheng2024comt} & 2024 & VQA & Common & T, I & MC & 3,853 \\
OmniBench \cite{li2024omnibench} & 2024 & VideoQA & Common & T, I, A & MC & 1,142 \\
\rowcolor{tablegrey} WorldQA \cite{zhang2024worldqa} & 2024 & VideoQA & Common & T, V, A & Open & 1,007 \\
MiCEval \cite{zhou2024miceval} & 2024 & VQA & Common & T, I & Open & 643 \\
\rowcolor{tablegrey} OlympiadBench \cite{he2024olympiadbench} & 2024 & ScienceQA & Maths, Physics & T, I & Open & 8,476 \\
MME-CoT \cite{jiang2025mmecot} & 2025 & VQA & Science, Math, Common & T, I & MC and Open & 1,130 \\
\rowcolor{tablegrey} EMMA \cite{hao2025emma} & 2025 & VQA & Science & T, I & MC and Open & 2,788 \\ 
VisualProcessBench \cite{wang2025visualprm} & 2025 & ScienceQA & Math, Science & T, I & MC and Open & 2,866 \\ 
\hline
\end{tabular}%
\vspace{1mm}
\caption{Datasets and Benchmarks for MCoT Training and Evaluation. ``MC'' and ``Open'' refer to multiple-choice and open-ended answer formats, while ``T'', ``I'', ``V'', and ``A'' represent Text, Image, Video, and Audio, respectively.
}
\label{tab:datasets}
\end{table}

\section{MCoT Datasets and Benchmarks}
\label{MCoT Datasets and Benchmarks}

MCoT reasoning necessitates specialized datasets and benchmarks to support both model finetuning and performance evaluation. 
Incorporating with \cref{tab:datasets}, this section surveys the landscape of MCoT-related resources, categorized into two key areas: datasets designed for finetuning MLLMs with reasoning rationales, and benchmarks developed to assess downstream capabilities, with or without accompanying rationales. 
These resources collectively address the diverse needs of training and evaluating MLLMs across multiple domains, modalities, and reasoning complexities. 

\subsection{Datasets for MLLMs Finetuning with Rationale}
Several studies explore to activate the MCoT reasoning capabilities of MLLMs through specific datasets. 
ScienceQA \cite{lu2022scienceqa} presents multimodal science questions paired with annotated answers, lectures, and explanations, illustrating how language models can leverage MCoT to enhance multi-hop reasoning. A-OKVQA \cite{schwenk2022okvqa} offers essential data for VQA by providing MLLMs with extensive commonsense and world knowledge. Building on ScienceQA, T-SciQ \cite{wang2024t} enriches reasoning rationale through advanced LLMs. VideoCoT \cite{wang2024videocot} supplies reasoning data for step-by-step video question answering (VideoQA), although its reasoning is limited to textual explanations. In contrast, VideoEspresso \cite{han2024videoespresso} provides VideoQA pairs that maintain spatial and temporal coherence along with multimodal reasoning annotations. Additionally, the MAVIS \cite{zhang2024mavis} dataset propels the training and application of MLLMs in mathematics by automating the generation of mathematical visual data, thus offering a wealth of aligned vision-language pairs and reasoning rationales.
EgoCoT \cite{mu2023embodiedgpt} and EMMA-X \cite{sun2024emmax} propose an egocentric dataset for the training model to execute the embodied tasks with sub-goal tasks. 
M3CoT \cite{Chen0ZC0C24} further advances VLLMs reasoning with multi-domain and multi-hop reasoning samples.
Meanwhile, MAmmoTH-VL-Instruct \cite{guo2024mammoth} constructs 12 million long-MCoT reasoning data across 118 datasets and 10 categories, advancing the long-MCoT reasoning capabilities of MLLMs. 
In addition, datasets such as LLaVA-CoT-100k \cite{xu2024llava_cot}, Mulberry-260k \cite{yao2024mulberry}, MM-Verify \cite{sun2025mmverify} and VisualPRM400K \cite{wang2025visualprm} are commonly proposed by their corresponding reasoning models to active the long-MCoT reasoning abilities. These datasets are commonly linked with learning-based rationale construction as we mentioned in \cref{sec:rationale_construction}.

\subsection{Benchmarks for Downstream Capability Assessment}

A variety of benchmarks have been developed to evaluate downstream capabilities, particularly in the domains of commonsense and scientific reasoning. As illustrated in \cref{tab:performance}, we present a performance comparison of MLLMs from various institutions across four benchmarks: MMMU \cite{yue2024mmmu}, MathVista \cite{lu2023mathvista}, Math-Vision \cite{wang2025mathvision}, and EMMA \cite{hao2025emma}. 
While MMMU and EMMA focus on multidisciplinary scenarios, MathVista and Math-Vision primarily assess mathematical reasoning.

\paragraph{Datasets without Rationale.} 
Several multimodal evaluation benchmarks have been widely adopted to assess the performance of MLLMs. Although these benchmarks do not provide rationales, their diversity and challenges suggest that, with the help of MCoT, MLLMs could further enhance their performance on these benchmarks. MMMU \cite{yue2024mmmu} involves visual-language questions across six core disciplines, aiming to measure the three fundamental abilities of perception, knowledge, and reasoning in LMMs. SEED \cite{li2024seed} further introduces the video modality to assess the understanding and generation capabilities of MLLMs. MathVista \cite{lu2023mathvista}, MathVerse \cite{zhang2024mathverse} and Math-Vision \cite{wang2025mathvision} specifically focuses on visual perception and reasoning in the mathematical domain. Emma \cite{hao2025emma} introduces reasoning challenges that cannot be solved by independent reasoning within each modality in solo, providing a comprehensive evaluation of multimodal reasoning abilities in MLLMs. 

In addition to the general and mathematical domains, MCoT demonstrates its effectiveness in various downstream tasks due to its step-by-step reasoning capabilities. Migician \cite{li2025migician} provides an evaluation of multi-image grounding and shows the effectiveness of MCoT in such tasks. RefAVS \cite{wang2024refavs} introduces grounding into the audiovisual context, incorporating temporal and spatial information, which can be further addressed through MCoT. VSIBench \cite{yang2024vsibench} offers an evaluation of spatial reasoning abilities in MLLMs. MeViS \cite{ding2023mevis} provides an evaluation of video segmentation with motion expressions. In particular, through the step-by-step reasoning of MCoT, hallucination phenomena that arise in MLLMs are expected to be further addressed. HallusionBench \cite{guan2024hallusionbench} evaluates hallucination phenomena in VLLMs, while AVTrustBench \cite{chowdhury2025avtrustbench} and AVHBench \cite{sung2024avhbench} assess hallucinations in audiovisual contexts, which can be further mitigated through MCoT reasoning. OSWorld \cite{xie2024osworld} and AgentClinic \cite{schmidgall2024agentclinic} provide benchmarks for assessing agent capabilities in multimodal scenarios, which can be enhanced through reasoning.

\paragraph{Datasets with Rationale.} 
With the emergence of OpenAI o1 \cite{jaech2024o1} and Deepseek-r1 \cite{guo2025deepseekr1}, interest in scaling test-time computation and slow-thinking has steadily grown, leading to the development of evaluation benchmarks designed to assess the quality of rationales generated by MLLMs during reasoning.
CoMT \cite{cheng2024comt} is introduced to address the limitations of traditional multimodal benchmarks that only reasoning with language, by requiring both multimodal input and output, aiming to better mimic human-like reasoning and explore complex visual operations.
WorldQA \cite{zhang2024worldqa} challenges MLLMs to answer questions using language, vision, and audio, while incorporating long-MCoT reasoning and world knowledge.
MiCEval \cite{zhou2024miceval} is crafted to evaluate the accuracy of reasoning chains by meticulously assessing the quality of both the descriptive component and each individual reasoning step.
OlympiadBench \cite{he2024olympiadbench} features 8000+ bilingual Olympiad-level mathematics and physics problems with annotated rationales to effectively evaluate the advanced capabilities of MLLMs.
MME-CoT \cite{jiang2025mmecot} provides a systematic evaluation of MCoT reasoning, revealing critical insights, including how reflection mechanisms enhance reasoning quality and how MCoT prompting can negatively impact perception-intensive tasks, potentially due to overthinking.
OmniBench \cite{li2024omnibench} is the first comprehensive evaluation benchmark involving text, vision, and audio.
 
\begin{table}[!t]
\fontsize{8.5}{9}\selectfont
\centering
\setlength{\tabcolsep}{1.6mm}
\begin{tabular}{lccccc}
\toprule
\rowcolor{headcolor}
\bf Model & \bf Params (B) & \bf MMMU (Val) & \bf MathVista (mini) & \bf Math-Vision & \bf EMMA (mini) \\ \midrule
Human & - & 88.6 & 60.3 & 68.82 & 77.75 \\
\rowcolor{tablegrey} Random Choice & - & 22.1 & 17.9 & 7.17 & 22.75 \\ \hline
\multicolumn{6}{c}{\textbf{\textit{OpenAI}}} \\
\rowcolor{tablegrey} o1 \cite{jaech2024o1} & - & 78.2 & 73.9 & - & 45.75 \\
GPT-4.5 \cite{gpt-4-5-system-card} & - & 74.4 & - & - & - \\
GPT-4o \cite{hurst2024gpt4o} & - & 69.1 & 63.8 & 30.39 & 36.00 \\
\rowcolor{tablegrey} GPT-4o mini \cite{hurst2024gpt4o} & - & 59.4 & 56.7 & - & - \\

\rowcolor{tablegrey} GPT-4V \cite{gpt-4v-system-card} & - & 56.8 & 49.9 & 23.98 & - \\ \hline
\multicolumn{6}{c}{\textbf{\textit{Google \& DeepMind}}} \\
\rowcolor{tablegrey} Gemini 2.0 Pro \cite{Gemini2} & - & 72.7 & - & - & - \\
Gemini 2.0 Flash \cite{Gemini2} & - & 71.7 & - & 41.3 & 48.00 \\
\rowcolor{tablegrey} Gemini 1.5 Pro \cite{team2024gemini1_5} & - & 65.8 & 63.9 & 19.24 & - \\ \hline
\multicolumn{6}{c}{\textbf{\textit{Anthropic}}} \\
\rowcolor{tablegrey} Claude 3.7 Sonnet \cite{Anthropic2024claude3} & - & 75 & - & - & 56.50 \\
Claude 3.5 Sonnet \cite{Anthropic2024claude3} & - & 70.4 & 67.7 & 37.99 & 37.00 \\
\rowcolor{tablegrey} Claude 3 Opus \cite{Anthropic2024claude3} & - & 59.4 & 50.5 & 27.13 & - \\
Claude 3 Sonnet \cite{Anthropic2024claude3} & - & 53.1 & 47.9 & - & - \\ \hline
\multicolumn{6}{c}{\textbf{\textit{xAI}}} \\
\rowcolor{tablegrey} Grok-3 \cite{grok3} & - & 78.0 & - & - & - \\
Grok-2 \cite{grok2} & - & 66.1 & 69.0 & - & - \\
\rowcolor{tablegrey} Grok-2 mini \cite{grok2} & - & 63.2 & 68.1 & - & - \\ \hline
\multicolumn{6}{c}{\textbf{\textit{Moonshot}}} \\
\rowcolor{tablegrey} Kimi-k1.5 \cite{team2025kimi1_5} & - & 70 & 74.9 (test) & 38.6 & 33.75 \\ \hline
\multicolumn{6}{c}{\textbf{\textit{Alibaba}}} \\
\rowcolor{tablegrey} QVQ-72B-Preview \cite{qwen2024qvq} & 72 & 70.3 & 71.4 & 35.9 & 32.00 \\
Qwen2.5-VL-72B \cite{bai2025qwen2_5_vl} & 72 & 70.2 & 74.8 & 38.1 & - \\
\rowcolor{tablegrey}Qwen2-VL-72B \cite{Qwen2-VL-abs-2409-12191} & 72 & 64.5 & 70.5 & 25.9 & 37.25 \\
Qwen2.5-VL-7B \cite{bai2025qwen2_5_vl} & 7 & 58.6 & 68.2 & 25.1 & - \\
\rowcolor{tablegrey} Qwen2-VL-7B \cite{Qwen2-VL-abs-2409-12191} & 7 & - & - & 16.3 & - \\ \hline
\multicolumn{6}{c}{\textbf{\textit{OpenGVLab}}} \\
\rowcolor{tablegrey} InternVL2.5 \cite{chen2024internvl_2_5} & 78 & 70.1 & - & - & 35.25 \\
InternVL2 \cite{chen2024internvl} & 76 & 58.2 & 65.5 & - & - \\ \hline
\multicolumn{6}{c}{\textbf{\textit{LLaMA}}} \\
\rowcolor{tablegrey} Llama-3.2-90B \cite{dubey2024llama3} & 90 & 60.3 & 57.3 & - & - \\
Llama-3.2-11B \cite{dubey2024llama3} & 11 & - & 48.6 & - & - \\ \hline
\multicolumn{6}{c}{\textbf{\textit{LLaVA}}} \\
\rowcolor{tablegrey} LLaVA-OneVision \cite{li2024llavaonevision} & 72 & 56.8 & 67.5 & - & 27.25 \\
LlaVA-NEXT-72B \cite{li2024llavanext} & 72 & 49.9 & 46.6 & - & - \\
\rowcolor{tablegrey} LLaVA-NEXT-34B \cite{li2024llavanext} & 34 & 48.1 & 46.5 & - & - \\
LLaVA-NEXT-8B \cite{li2024llavanext} & 8 & 41.7 & 37.5 & - & - \\
\rowcolor{tablegrey} LLaVA-Reasoner \cite{zhang2024llavareasoner} & 8 & 40.0 & 50.6 & - & - \\
LLaVA-1.5 \cite{liu2024llava} & 13 & 36.4 & 27.6 & 11.12 & - \\ 

\hline
\multicolumn{6}{c}{\textbf{\textit{Community}}} \\
\rowcolor{tablegrey} Mulberry \cite{yao2024mulberry} & 7 & 55.0 & 63.1 & - & - \\
MAmmoTH-VL \cite{guo2024mammoth} & 8 & 50.8 & 67.6 & 24.4 & - \\

\rowcolor{tablegrey} MM-Eureka \cite{meng2025mmeureka} & 8 & - & 67.1 & 22.2 & - \\
MM-Eureka-Zero \cite{meng2025mmeureka} & 38 & - & 64.2 & 26.6 & - \\
\rowcolor{tablegrey} Curr-ReFT \cite{deng2025vlmreasoning} & 7 & - & 64.5 & - & - \\
Curr-ReFT \cite{deng2025vlmreasoning} & 3 & - & 58.6 & - & - \\
\rowcolor{tablegrey} LMM-R1 \cite{peng2025lmmr1} & 3 & - & 63.2 & 26.35 & -  \\ 
LlamaV-o1 \cite{thawakar2025llamav_o1} & 11 & - & 54.4 & - & - \\
\rowcolor{tablegrey} R1-Onevision \cite{R1-OneVision} & 7 & - & - & 26.16 & - \\
Virgo \cite{du2025virgo} & 7 & 46.7 & - & 24.0 & - \\
\rowcolor{tablegrey} Insight-V \cite{dong2024insight} & 8 & 42.0 & 49.8 & - & - \\ 
R1-VL \cite{zhang2025r1vl} & 7  & 63.5 & 24.7 & - & - \\


\hline
\end{tabular}%
\vspace{1mm}
\caption{Performance comparison of MLLMs from various institutions across four benchmarks: MMMU (Val), MathVista (Mini), Math-Vision, and EMMA (Mini).
}
\label{tab:performance}
\end{table}

\section{Limitations, Challenges and Future Directions}
\label{Limitations}

Despite the increasing attention and research efforts devoted to MCoT, several critical aspects remain unresolved and underexplored, which might be the key bottlenecks for achieving human-level multimodal AGI. 
Below, we summarize some challenges to shed light on future works.

\paragraph{Computational Sustainability and Slow-thinking Paradox.} 
Despite significant advancements, the reliance on extensive test-time scaling and slow-thinking \cite{li2025system,gan2025rethinking,chen2025towards} to support long-MCoT reasoning poses substantial challenges. 
The exponential growth in computational resources and training data required to sustain deep reasoning processes remains a critical bottleneck, necessitating innovations in algorithm efficiency like RL \cite{guo2025deepseekr1} and hardware acceleration.

\paragraph{Lack of Reasoning in General Scenarios.} 
The existing long-MCoT framework primarily focuses on verifiable data in mathematics and science, yet it lacks robust reasoning capabilities for general scenarios. Tasks in mathematics and science, characterized by their strict logical structure and unique solutions, have been extensively explored for test-time scaling using ORM \cite{multimodal-open-r1,meng2025mmeureka} and PRM \cite{liu2024diving,wang2025visualprm}. However, in general scenarios, answers are rarely fixed, often encompassing multiple plausible interpretations and inferences. This variability renders reasoning frameworks rooted in mathematics and science less effective. 
The deficiency in reasoning manifests as an inability to adequately evaluate the inference process when models address tasks in general contexts involving complex situations, ambiguity, and multifactorial influences. 
Future research should explore the open-ended reward models to achieve robust long-chain reasoning in multimodal general scenarios.

\paragraph{Error Propagation in Extended Reasoning Chains.} 
A major concern in long-chain reasoning in current MCoT systems is the error snowballing effect \cite{gan2025rethinking}, where small inaccuracies in early steps can amplify through subsequent stages and lead to catastrophic outcomes. 
Traditional confidence calibration techniques fail to address the unique challenges of multimodal error propagation where different modalities may contradict each other while maintaining high self-consistency scores. 
Developing quantitative metrics to diagnose, quantify, and mitigate these cumulative errors is an unresolved issue that requires rigorous investigation.

\paragraph{Symbolic-Neural Integration Gap.} 
While neural models excel at pattern recognition, their inability to perform rigorous symbolic operations undermines complex reasoning. 
Hybrid neurosymbolic architectures \cite{amizadeh2020neuro,xu2024faithful,delong2024neurosymbolic} can help achieve improvement on mathematical proofs, but might still suffer from knowledge grounding issues. 
The fundamental challenge lies in developing seamless interfaces between distributed representations and discrete symbolic systems, particularly for cross-modal symbolic manipulation (\eg, converting geometric diagrams to formal proofs).

\paragraph{Dynamic Environment Adaptation and Adaptive Chain Length.} 
Most MCoT systems assume static input conditions, severely limiting real-world applicability. 
The ``frozen reasoning'' paradox emerges when processing streaming multimodal inputs—most current architectures cannot revise earlier conclusions upon receiving new evidence without restarting the entire chain. 
Also, the development of resource-efficient reasoning frameworks that can dynamically adjust chain length or the number of reasoning steps \cite{dong2024insight} in response to computational constraints is critical. 
Adaptive strategies based on real-time evaluation and feedback mechanisms will be key to balancing reasoning accuracy with resource efficiency.

\paragraph{Hallucination Prevention.} 
A critical challenge in existing MCoT reasoning can be mitigating hallucinations \cite{zheng2024thinking}, where models produce plausible yet factually incorrect or inconsistent outputs. 
This issue undermines the reliability of the reasoning process and is especially pronounced in multimodal settings, where integrating diverse data sources can lead to misaligned contexts and spurious information. 
Future work should focus on robust cross-modal alignment methods and verification mechanisms at each reasoning step. Techniques such as uncertainty quantification, adversarial training, and leveraging external knowledge bases have shown promise in reducing hallucinations. Additionally, insights from human error detection may inspire novel strategies to enhance both the accuracy and interpretability of multimodal reasoning systems.

\paragraph{Data Selection, Annotation, and Augmentation Strategies.} 
Recent studies have demonstrated that carefully curated datasets can activate long-MCoT reasoning capabilities in models \cite{xu2024llava_cot,yao2024mulberry,ye2025limo}. 
However, automating the selection and annotation of data suitable for extended reasoning remains an open challenge. Leveraging semi-supervised or self-supervised learning strategies, alongside RL approaches, could reduce the reliance on extensive manual annotations.

\paragraph{Modality Imbalance and High-Dimensional Modal Integration.} 
Current research indicates that progress across different modalities is uneven, with some modalities, such as text and images, advancing more rapidly than others. Future work should address the integration of higher-dimensional modalities (\eg, 3D data, sensor information) to achieve a more balanced and comprehensive multimodal reasoning framework that leverages the unique characteristics of each modality.

\paragraph{Interdisciplinary Integration with Cognitive Science.} 
The inherent complexity of MCoT reasoning calls for an interdisciplinary approach that integrates insights from cognitive science, psychology, and neuroscience. 
Drawing from human decision-making and cognitive theories \cite{wu2023role,dai2024interpretable,wu2024mind} can inspire novel reasoning architectures that more closely mimic human thought processes, thereby enhancing both performance and interpretability.

\paragraph{Embodied Reasoning Limitations.} 
Most MCoT systems operate in abstract symbol spaces disconnected from physical embodiment. 
Closing this simulation-to-reality gap necessitates tight integration of proprioceptive feedback, haptic understanding, and dynamic world modeling—challenges that current architecture designs barely address.

\paragraph{Explainable Reasoning and Theoretical Support.} 
As models become increasingly complex, the interpretability of decision-making processes is essential for building trust and enabling practical deployment. 
Although CoT or MCoT reasoning provides intermediate steps, its underlying theoretical mechanisms remain largely opaque, leaving the model as a ``black box'' that simply produces predictions and outputs. 
Developing methodologies that offer transparent, traceable reasoning paths will not only enhance model explainability but also facilitate debugging and further optimization of MCoT reasoning systems. 
Strengthening theoretical support is therefore crucial to achieving truly explainable reasoning.

\paragraph{Ethical, Robustness and Safety Reasoning.} 
As MCoT systems become more powerful, ensuring AI safety and robustness against adversarial perturbations is paramount. 
Integrating more robust reasoning techniques and approaches may enhance system transparency and provide a safety net against potential failures. 
Also, developing multimodal constitutional AI frameworks that can parse and apply ethical constraints across modalities becomes crucial as these systems approach real-world deployment.
Further research is needed to quantify and mitigate risks associated with adversarial attacks and other safety concerns in multimodal reasoning environments.

\section{Conclusion}
This survey presents the first systematic review of multimodal chain-of-thought (MCoT) reasoning.  
We start by providing definitions and elucidated foundational concepts, laying the groundwork for methodologies.  
A comprehensive taxonomy categorizes diverse approaches and reasoning paradigms under various perspectives to tackle challenges unique to modalities such as image, video, speech, audio, 3D, and structured data.  
We consolidate comprehensive MCoT-related datasets and benchmarks to provide an accurate overview of the current resource landscape.  
Furthermore, our review examines relevant applications, highlighting successes in important domains including robotics, healthcare, autonomous driving, social science and multimodal generation.  
Lastly, we outline promising future research directions that aim to overcome these limitations and drive progress toward multimodal AGI.  
We openly share all related resources and information on MCoT to facilitate follow-up research in this rapidly evolving domain.

\bibliography{main}
\bibliographystyle{unsrtnat}


\end{CJK}
\end{document}